\documentclass[11pt]{article}

\usepackage[preprint]{acl}

\usepackage{times}
\usepackage{latexsym}

\usepackage[T1]{fontenc}

\usepackage{tabularx}
\usepackage{booktabs}

\usepackage{longtable}

\usepackage{amsmath,amssymb}
\usepackage{graphicx}
\usepackage{tikz}
\usetikzlibrary{positioning,arrows.meta}

\tikzset{
  exblock/.style={
    draw, rounded corners=1pt, align=center, inner sep=5pt,
    font=\small
  },
  exloss/.style={exblock, dashed},
  exflow/.style={-{Latex[length=2mm]}, semithick},
  extrain/.style={exflow, dashed}
}

\usepackage[utf8]{inputenc}

\usepackage{microtype}

\usepackage{inconsolata}

\usepackage{graphicx}

\title{Predicting Training Exposure and Response Reliability
in a Small Language Model with a Two-Expert Architecture}

\author{Artem Suslov \\
\texttt{artem.suslov.227@gmail.com}}

\begin{document}
\onecolumn
\begingroup
\renewcommand{\twocolumn}[1][]{#1}
\maketitle
\endgroup

\begin{abstract}
Can a language model distinguish information encountered during
training from information it cannot reliably reproduce?
This study investigates this question using controlled fine-tuning of
Qwen3-0.6B on synthetic biographies. A Knowledge Expert constructs
a question-conditioned vector and predicts whether the requested
information was shown, unshown, or mixed during fine-tuning.
A Decision Expert combines this vector with question and response
hidden states to assess response reliability. The hidden-state
two-expert configurations achieved 73.86--74.56\% macro-F1,
compared with 53.48\% for a text-based baseline.
Exposure prediction was less successful, particularly for
mixed-exposure questions. Explicit model-weight input provided
modest, uneven gains, while asymmetric loss changed the
distribution of errors. These results provide preliminary
evidence for joint exposure prediction and response assessment
through auxiliary experts.
\paragraph{Key findings}
\begin{itemize}
    \item A two-expert architecture using a question-conditioned
    knowledge vector achieved 73.86--74.56\% macro-F1 in
    three-class response assessment, compared with 53.48\%
    for the text-based baseline.

    \item Controlled biography fine-tuning enabled separate
    evaluation of factual exposure and response reliability.
    Exposure classification was less accurate than response
    assessment, with overall accuracy of 67.10--68.55\%
    across supervised configurations.

    \item Mixed exposure was harder to identify than uniformly
    shown or unshown information: the hidden-state Knowledge
    Expert achieved 49.5--52.5\% accuracy on mixed cases,
    compared with 69.5--74.5\% on shown or unshown cases.

    \item Explicit model-weight input produced modest,
    uneven improvements in response assessment and no
    improvement in overall exposure-classification macro-F1.

    \item Asymmetric loss shifted the distribution of decision
    errors: without model-weight input, accuracy increased
    on unshown and mixed cases while decreasing on shown
    cases, despite similar overall performance.
\end{itemize}
\end{abstract}

\section{Introduction}
\label{sec:introduction}

Large language models (LLMs) can acquire and reproduce substantial
amounts of factual information, yet they remain susceptible to
generating incorrect, unsupported, or partially accurate responses.
These errors may arise from limitations in knowledge acquisition,
failures to retrieve previously learned information, or the
generation of plausible but factually incorrect statements.
The presence of a factual association in a model's
training data does not guarantee its successful acquisition or
reliable reproduction. This distinction raises the question of
whether the internal representations of an LLM can robustly
assess the availability of factual knowledge and the reliability
of its generated responses.

Existing approaches to knowledge-awareness assessment commonly
estimate response correctness using output probabilities
\citep{kadavath2022languagemodelsmostlyknow},
repeated generations
\citep{farquhar2024detecting},
or classifiers trained on hidden states
\citep{azaria-mitchell-2023-internal,orgad2025llmsknowshowintrinsic,cencerrado2026answerneededpredictingllm}.
However, predicting whether an answer is correct is not
necessarily equivalent to determining whether the model
possesses the relevant factual knowledge. A model may generate
an incorrect response despite having encountered the correct
fact during training. Conversely, it may produce a correct
answer without reliably retrieving the corresponding factual
association. Previous studies have demonstrated that apparently
correct answers may be supported by incorrect intermediate
reasoning \citep{ishii-etal-2025-fine}. Therefore, evaluating
knowledge awareness requires distinguishing factual correctness
from the availability and successful retrieval of acquired
knowledge.

This distinction becomes particularly important when a model
is trained on closely related factual associations. For
example, learning the residence, occupation, or birthplace of
multiple individuals requires distinguishing similar entities
and attributes. Such knowledge may be particularly challenging
to retrieve when several associations share similar linguistic
structures. Moreover, evaluating factual knowledge using
general question-answering benchmarks does not necessarily
establish whether the underlying facts were encountered during
pretraining and training. Controlled fine-tuning on deliberately 
constructed
factual associations offers an alternative setting in which
the exposure of a model to specific information can be
documented and its subsequent reproduction evaluated.

The present study investigates a two-expert architecture for
assessing factual knowledge awareness in a fine-tuned
Qwen3-0.6B model \citep{qwen3technicalreport}. The proposed framework 
separates the
construction of a knowledge representation from the
classification of generated responses. The first expert,
referred to as the Knowledge Expert, produces a
128-dimensional, question-conditioned knowledge vector
intended to represent the model's knowledge relevant to
the input question. This representation is constructed from
internal model states, with additional configurations
incorporating information derived from the model's weights.
The second expert, referred to as the Decision Expert,
evaluates the model's response using its internal
representations and, depending on the experimental
configuration, the knowledge vector.

Rather than limiting the evaluation to binary answer
correctness, the proposed approach distinguishes three
response categories: \emph{reliable}, \emph{questionable},
and \emph{no training evidence}. These operational labels
combine response assessment with evidence of fine-tuning
exposure; absence of training evidence does not necessarily
imply that a response is factually incorrect. In this study, these categories
serve as operational indicators of the model's ability to
reproduce factual knowledge. The distinction between
partial and complete factual errors allows us to examine
whether an auxiliary classifier can identify varying
degrees of factual reliability.

To investigate the effectiveness of this architecture,
this study constructs a controlled dataset of factual associations,
including closely related and potentially confusable facts.
These associations are introduced through fine-tuning,
allowing us to distinguish facts included in the fine-tuning
corpus from those deliberately excluded. We then evaluate
whether the model's internal representations and the
additional knowledge vector can predict the factual
correctness of its responses.

The study addresses the following research questions:

\begin{itemize}
    \item \textbf{RQ1:} To what extent can a two-expert
    architecture using a question-conditioned knowledge vector
    predict fine-tuning exposure and classify response reliability
    in a small language model, including questions that combine
    shown and unshown information?

    \item \textbf{RQ2:} How does the hidden-state two-expert
    architecture compare with a text-based Decision Expert
    baseline, and how does explicit access to model-weight
    information affect exposure prediction and response
    assessment?

    \item \textbf{RQ3:} How do symmetric and asymmetric
    training objectives affect classification performance
    and the distribution of errors, relative to fixed-parameter
    random-loss controls?
\end{itemize}

To address these questions, the study makes three
contributions. First, we construct a controlled synthetic
biography dataset that distinguishes shown, unshown, and
mixed fine-tuning exposure, enabling separate evaluation
of factual recall, exposure prediction, and response
reliability. Second, we introduce and evaluate a two-expert
architecture in which a question-conditioned knowledge
vector supports both exposure prediction and response-level
assessment. Third, we compare text-based, hidden-state,
and weight-augmented configurations, together with symmetric
and asymmetric objectives and fixed-parameter random-loss
controls. The results demonstrate the feasibility of the
two-expert approach in this controlled setting. Explicit
model-weight input yields modest, uneven gains in response
assessment without improving overall exposure-classification
macro-F1, while asymmetric loss changes the distribution
of decision errors.

The remainder of this paper is organized as follows.
Section~\ref{sec:litreview} reviews research on knowledge
awareness, factual knowledge acquisition, and internal
representations.
Section~\ref{sec:methods} describes the controlled biography
dataset, language-model fine-tuning, and expert architectures
and training objectives.
Section~\ref{sec:experiment_setup} presents the experimental
configurations and evaluation procedure.
Section~\ref{sec:results} reports exposure-prediction and
response-assessment results.
Section~\ref{sec:discussion} discusses their implications,
limitations, and directions for further work.
The appendices provide dataset specifications, training
details, and additional evaluation results.

\section{Evaluating Knowledge and Knowledge Awareness in Language Models}
\label{sec:litreview}

\subsection{Knowledge Awareness and Correctness Prediction in Language Models}

Research on knowledge awareness in LLMs investigates whether models can 
distinguish questions they are capable of answering from those beyond their knowledge boundaries. 
However, knowledge possession, awareness of knowledge, and response correctness are not 
equivalent concepts \citep{li2025knowledgeboundarylargelanguage}. Existing approaches typically operationalize 
knowledge awareness through the correctness of generated answers, using output 
probabilities \citep{kadavath2022languagemodelsmostlyknow,chen2026queryleveluncertaintylargelanguage}, 
multiple generations \citep{farquhar2024detecting}, 
or internal representations \citep{azaria-mitchell-2023-internal,orgad2025llmsknowshowintrinsic,ni-etal-2025-towards,cencerrado2026answerneededpredictingllm}.

\citet{kadavath2022languagemodelsmostlyknow} introduced two measures: $(P(\mathrm{True}))$, 
estimating the correctness of a proposed answer, and $(P(\mathrm{IK}))$, 
predicting whether a model can answer a question correctly before generation. 
The latter employs an additional prediction head trained on empirical 
answering performance. Other approaches estimate uncertainty from output 
distributions or repeated generations. \citet{farquhar2024detecting} proposed 
semantic entropy, measuring uncertainty across semantically distinct 
responses, while \citet{kossen2024semanticentropyprobesrobust} trained probes to approximate 
this uncertainty directly from hidden states. \citet{chen2026queryleveluncertaintylargelanguage} subsequently 
proposed a training-free approach aggregating internal confidence estimates 
across layers and tokens before response generation.

A related line of research examines whether factual correctness can be 
recovered from internal representations.
 \citet{azaria-mitchell-2023-internal}
trained classifiers on hidden states to distinguish true from false 
statements, reporting 71–83\% accuracy. 
\citet{burns2024discoveringlatentknowledgelanguage} proposed 
Contrast-Consistent Search, which identifies latent truthfulness signals 
without labelled training examples, 
although \citet{farquhar2024detecting}
subsequently demonstrated limitations of this unsupervised approach. 
\citet{marks2024geometrytruthemergentlinear} identified approximately linear directions 
corresponding to truthfulness in hidden-state space.

More recent research has focused on detecting model-specific knowledge 
boundaries. 
\citet{orgad2025llmsknowshowintrinsic} showed that internal representations 
can contain information about factual correctness not reflected in 
generated responses. 
\citet{gekhman2025insideouthiddenfactualknowledge} further distinguished 
externally expressed knowledge from internally recoverable factual 
knowledge, demonstrating that models may encode correct answers 
they fail to generate. 
\citet{ni-etal-2025-towards} investigated knowledge-boundary 
prediction using hidden states extracted before and after generation, 
while 
\citet{cencerrado2026answerneededpredictingllm} demonstrated that question-only 
linear probes can predict subsequent answer correctness without 
generating a response. 
\citet{ashuach-etal-2026-masked} examined whether such 
predictions reflect model-specific knowledge rather than general 
question difficulty. Other methods include gradient-subspace 
analysis \citep{wang2026gradeprobingknowledgegaps}, compact knowledge-boundary 
subspaces \citep{huang-etal-2026-uncovering} , and comparisons between 
internal probes and model output logits \citep{villuri2026internalprobesbeatreading}.

These studies demonstrate that information 
predictive of factual correctness is accessible through 
internal representations. Nevertheless, predicting whether 
an answer is correct does not necessarily establish that a model 
recognizes the presence or absence of its own knowledge. Moreover, 
most approaches employ binary distinctions between correct and 
incorrect answers rather than explicitly representing partially 
correct responses.

\subsection{Sources and Representations of Factual Knowledge}
Another important distinction concerns the origin of evaluated knowledge. 
Most knowledge-awareness studies examine general factual knowledge through 
benchmarks such as TriviaQA \citep{2017arXivtriviaqa}, 
Natural Questions \citep{sentence_transformers2024natural_questions}, and other 
question-answering datasets. These include geographical, biographical, 
historical, and encyclopedic facts potentially acquired 
during pretraining. Although benchmark questions may be excluded 
from the training of auxiliary classifiers, the underlying factual 
information is not necessarily novel to the evaluated language models. 
Consequently, incorrect generation cannot be interpreted as definitive 
evidence that the corresponding knowledge is absent from the model's parameters.

Other studies address different forms of uncertainty. 
\citet{amayuelas-etal-2024-knowledge}, for example, investigated known-unknown 
questions that lack definitive answers, distinguishing uncertainty 
arising from the question itself from failures to retrieve factual 
knowledge. This differs from determining whether a model has 
successfully acquired a specific factual association.

Research on factual knowledge acquisition provides an alternative 
experimental framework. \citet{NEURIPS2024_6fdf57c7} investigated how 
language models acquire and retain factual information during 
pretraining. \citet{ou-etal-2025-llms} examined the evolution of knowledge 
circuits during continual pretraining. 
Similarly, \citet{zucchet2025languagemodelslearnfacts}
studied factual learning dynamics using synthetic knowledge. 
\citet{NEURIPS2025_90043ebd} investigated factual 
knowledge injection through fine-tuning and demonstrated that the 
format and characteristics of training examples affect subsequent 
knowledge retrieval and generalization.

These studies provide more direct control over the acquisition 
of specific factual associations. However, they primarily investigate 
whether knowledge can be learned and subsequently reproduced, rather 
than whether internal representations can identify which facts have 
been successfully acquired. Controlled fine-tuning therefore offers an 
opportunity to study knowledge-boundary classification while distinguishing 
newly introduced information from knowledge potentially acquired during pretraining.

\subsection{Knowledge Vectors and Internal Representations}
Research on the internal organization of factual knowledge has 
also investigated whether knowledge can be represented through 
individual neurons, directions in activation space, or compact 
vectors. 
\citet{dai-etal-2022-knowledge} introduced knowledge neurons, identifying 
feedforward neurons associated with particular factual associations 
through attribution analysis. 
\citet{meng2023locatingeditingfactualassociations} demonstrated that 
factual associations can be localized and modified through rank-one 
updates of transformer feedforward weights, conceptualizing factual 
retrieval in terms of key–value relationships.

Other approaches construct vectors from model activations. 
\citet{todd2024functionvectorslargelanguage} introduced function vectors, derived from 
attention-head activations associated with demonstrated input–output 
tasks. 
\citet{marks2024geometrytruthemergentlinear} constructed truthfulness directions 
from differences between hidden-state representations of true and false 
statements. 
\citet{ferrando2025iknowentityknowledge} employed sparse autoencoders to 
identify latent features associated with recognition of known and 
unknown entities. 
\citet{huang-etal-2026-uncovering} similarly investigated compact 
subspaces associated with knowledge-boundary prediction, 
while 
\citet{wang2026gradeprobingknowledgegaps} used gradient-subspace 
characteristics as features for detecting knowledge gaps.

These approaches differ substantially in their objectives. 
Knowledge neurons and model-editing methods investigate 
the localization and modification of factual associations, whereas 
function vectors represent task-related computations. Truthfulness 
directions and uncertainty-related features, by contrast, can be used 
to classify statements or estimate response reliability. However, a 
representation associated with factual storage or retrieval is not 
necessarily an effective representation for predicting whether a 
particular model can answer a given question.

Three considerations motivate the present study.
First, evaluations using general factual benchmarks such as
TriviaQA and Natural Questions concern information whose
exposure during language-model pretraining is often uncertain
\citep{kadavath2022languagemodelsmostlyknow,
orgad2025llmsknowshowintrinsic,
ni-etal-2025-towards,
cencerrado2026answerneededpredictingllm}.
These evaluations measure correctness and retrievability,
but do not directly establish whether particular facts were
encountered during training. Controlled factual fine-tuning
offers a complementary setting
\citep{NEURIPS2025_90043ebd}: recording which canonical
biographies are included provides explicit exposure labels.
Exposure nevertheless remains a proxy for knowledge, since
inclusion does not guarantee retention and excluding a
biography does not guarantee that all its component names
and values were absent from other training examples.

Second, training exposure and response reliability describe
different properties. Prior work already provides multicategory
knowledge characterization, including the five states of
\citet{xiao2025kscopeframeworkcharacterizingknowledge},
and fine-grained assessment of factual errors
\citep{ishii-etal-2025-fine}. The present study instead examines
the relationship between exposure prediction and response
assessment, including questions that combine shown and unshown
information. Mixed exposure is distinct from partial factual
correctness: information absent from fine-tuning can still
be true, while exposure does not guarantee a correct response.

Third, factual associations can be localized and modified
within transformer parameters
\citep{dai-etal-2022-knowledge,
meng2023locatingeditingfactualassociations}, but this does
not establish whether explicit access to weights improves
an auxiliary predictor that already receives hidden states.
Comparing otherwise corresponding architectures with and
without a question-conditioned weight reader therefore
provides an empirical test of its additional value.

These considerations motivate the joint investigation of
fine-tuning exposure and response reliability in a controlled
factual setting.

\section{Data and Methods}
\label{sec:methods}

\subsection{Fact Dataset Preparation and Learning}

Knowledge in this study concerns the model's ability to
reproduce diverse and potentially confusable information about a
set of homogeneous objects. This knowledge resembles a catalogue
or phone book containing structurally similar records referring
to distinct objects.

Each biography in the dataset contains natural-language
information about a person: their name, birth year, city of
residence, street address, year of moving into the current
residence, and housemate's name. Examples and
dataset details are provided in Appendix~\ref{app:dataset}.
All subsequent information-retrieval requests refer to
individual people, so the generation script requires
each person's name to be unique.

Question--answer pairs were generated from these biographies.
The dataset contains five groups of questions: retrieval of
a single biographical fact, retrieval of multiple biographical
facts, full-biography recall, verification of a true fact,
and verification of a false fact.

The Qwen3-0.6B model was fine-tuned on these question--answer
pairs. With approximately 600 million parameters, its size
is comparable to that of larger BERT models, while it can
generate coherent text. Its compact size makes inference
and fine-tuning relatively manageable on modest equipment,
making it a practical model for this experiment.

The model was trained on 155,844 question--answer pairs for
five epochs to memorize the biographies and reproduce their
information in response to related questions. The training
dataset distribution, training parameters, and evaluation
procedure are described in Appendix~\ref{app:model_training}.
By the end of the fifth epoch, the model correctly reproduced
all answers in the 384-question sample drawn from the training
data (Table~\ref{tab:training_performance}). This result
demonstrates complete recall on the sampled questions,
rather than establishing memorization of every fact in
the dataset.

After fine-tuning, the selected model was evaluated on a
separate post-training evaluation panel containing questions
about biographies included in fine-tuning and biographies
excluded from it, as described in
Appendix~\ref{app:post_train_validation}.

\subsection{Two-Stage Classification of Knowledge and Awareness}

Three architectural configurations were trained and tested.
The target configuration uses a two-expert architecture,
with both experts optimized jointly under the same loss
function. The Knowledge Expert receives the hidden state
of the question's last token, produces a 128-dimensional
knowledge vector, and predicts exposure: whether the
requested information concerns biographies shown during
fine-tuning, unshown biographies, or a mixture of both.
The Decision Expert receives the knowledge vector together
with the question and response hidden states.
Both two-expert configurations were evaluated with three
loss functions.

The baseline configuration contains only a Decision Expert
and uses text features derived from the question and
generated response. This configuration serves as the
baseline for comparison.

The target configuration contains two experts: the Knowledge
Expert produces the knowledge vector and predicts exposure,
while the Decision Expert classifies the response as
\emph{reliable}, \emph{questionable}, or \emph{no training
evidence}. This is the main configuration used to test
the proposed architecture.

The weight-augmented configuration extends the target
configuration. Its knowledge vector is produced using
both the question's last-token hidden state and explicit
access to the frozen model's weights, divided into
291,121 chunks.

Except for the baseline configuration, both the hidden-state-only
and weight-augmented configurations were evaluated using three
loss functions.

$e_i$ and $y_i$ denote the exposure and decision labels for
example i, and let \(\ell_i^{\mathrm{exp}}\) and
\(\ell_i^{\mathrm{dec}}\) denote the corresponding logits. The
per-example cross-entropy terms are
\[
K_i = \operatorname{CE}(\ell_i^{\mathrm{exp}}, e_i),
\qquad
D_i = \operatorname{CE}(\ell_i^{\mathrm{dec}}, y_i).
\]
Each objective is averaged over a minibatch of B examples.

\paragraph{Symmetric loss.}
The symmetric objective gives equal weight to the exposure and decision
cross-entropies and weights all examples uniformly:
\[
\mathcal{L}_{\mathrm{sym}}
= \frac{1}{B}\sum_{i=1}^{B}(K_i + D_i).
\]
The baseline configuration uses the mean decision cross-entropy
because it has a single decision head.

\paragraph{Asymmetric loss.}
The asymmetric objective scales the decision cross-entropy by response error
severity and penalizes assigning high probability to the reliable class
when the target belongs to another class. Let
\(p_i = \operatorname{softmax}(\ell_i^{\mathrm{dec}})_{\mathrm{reliable}}\).
The objective is
\[
\mathcal{L}_{\mathrm{asym}}
= \frac{1}{B}\sum_{i=1}^{B}
  \left[
    K_i + s_iD_i
    + a_i p_i\,\mathbb{1}\{y_i \ne \mathrm{reliable}\}
  \right],
\]
where the severity multiplier si and false-acceptance cost ai
are
\begin{align*}
s_i &= \min\{20,\ 1 + 0.25m_i + 3c_i + 5u_i\}, \\
a_i &= \min\{20,\ \phantom{1 + {}}0.15m_i + 2.5c_i + 4u_i\}.
\end{align*}
In the general formulation, $m_i$, $c_i$, and $u_i$ represent
counts of missing, contradicted, and unsupported claims. In the reported
experiments, lexical reference matching provided a proxy for error
severity. A matching response received $m_i=0$, and a mismatch received
$m_i$ equal to the number of required claims. The remaining counts were
set to $c_i=u_i=0$. 

\paragraph{Random loss.}
The random-loss control draws a fresh independent uniform value for
each example presentation:
\[
\mathcal{L}_{\mathrm{rand}}
= \frac{1}{B}\sum_{i=1}^{B} U_i,
\qquad
U_i \overset{\mathrm{iid}}{\sim} \operatorname{Uniform}(0,1).
\]
The draws are independent of the model outputs and labels.
The loss therefore provides no gradient to either expert,
and the training loop skips optimizer updates.

The purpose of introducing this deliberately uninformative
loss function is to test whether the supervised architectures
learn meaningful classification beyond the behavior of
their fixed-parameter counterparts.

\subsubsection{Baseline Configuration}

\begin{figure}[htbp]
\centering
\resizebox{!}{0.38\textheight}{%
\begin{tikzpicture}[node distance=4mm and 9mm]
  \node[exblock] (question) {Question text};
  \node[exblock, right=of question] (reply) {Generated response text};

  \path (question.south) -- (reply.south) coordinate[midway] (inputs);
  \node[exblock, below=7mm of inputs] (encoder) {
    \textbf{Shared text encoder}\\
    Hashed unigrams and bigrams\\
    Embedding table 16,38432\\
    Mean pool excluding padding
  };
  \node[exblock, below=of encoder] (concat) {
    Concatenate \(u_q,u_a\in\mathbb R^{32}\)\\
    \([u_q;u_a]\in\mathbb R^{64}\)
  };
  \node[exblock, below=of concat] (judge) {
    \textbf{Decision Expert (judge)}\\
    LayerNorm 64\\
    Linear 64256 + GELU\\
    Linear 256128 + GELU\\
    Linear 1283
  };
  \node[exblock, below=of judge] (output) {\textbf{Three reliability logits:} \\
  reliable\\
  questionable\\
  no training evidence};

  \draw[exflow] (question.south) |- ([yshift=3mm]encoder.north)
    -| ([xshift=-8mm]encoder.north);
  \draw[exflow] (reply.south) |- ([yshift=3mm]encoder.north)
    -| ([xshift=8mm]encoder.north);
  \draw[exflow] (encoder) -- (concat);
  \draw[exflow] (concat) -- (judge);
  \draw[exflow] (judge) -- (output);
\end{tikzpicture}%
}
\caption{Baseline configuration. A shared encoder processes the question and generated
response separately. The judge classifies their concatenated text embeddings.}
\label{fig:expert-text-pair}
\end{figure}
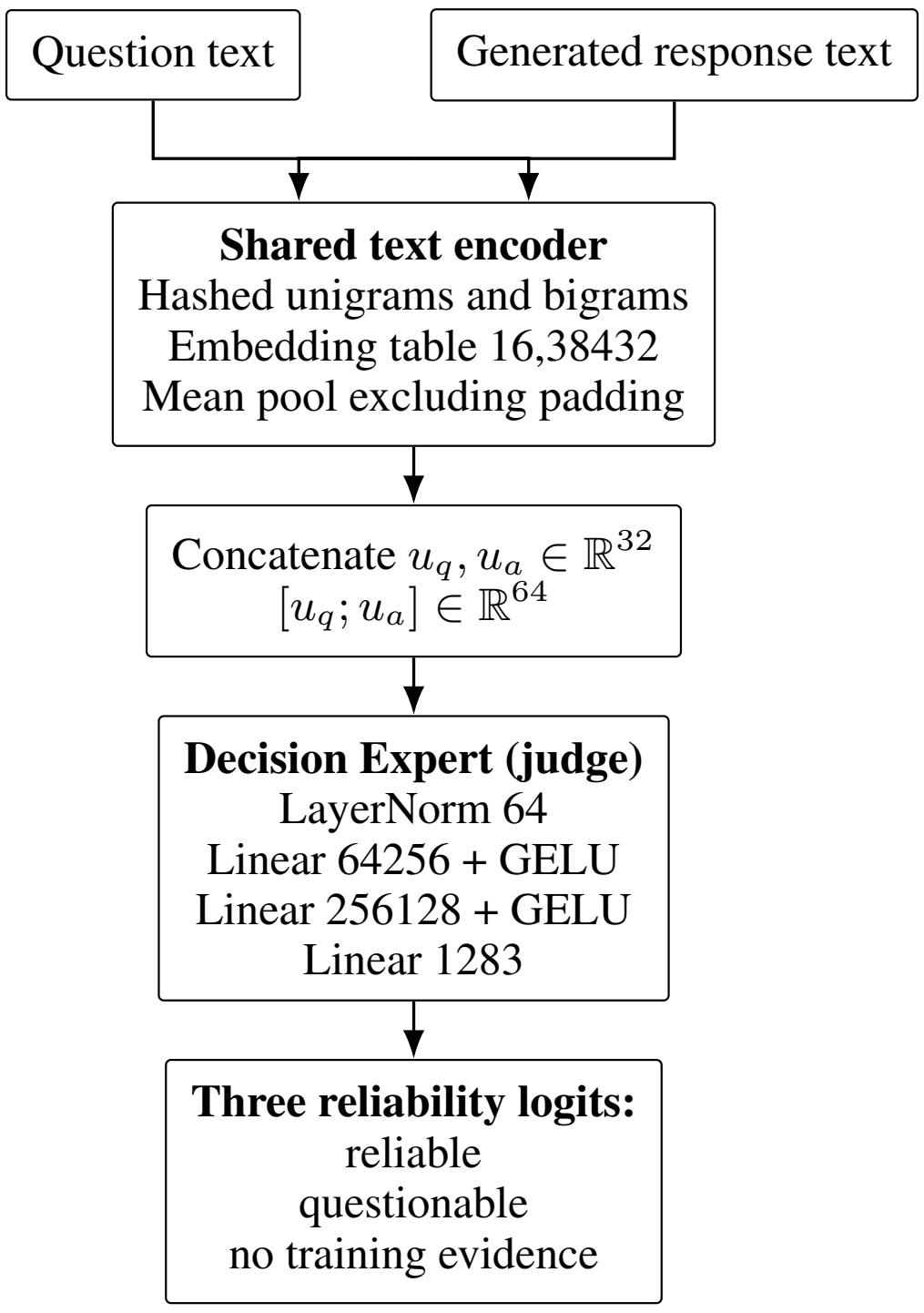

The Decision Expert receives only the question and the 
response generated by the model (Figure~\ref{fig:expert-text-pair}). It converts 
each text into hashed token 
unigrams and bigrams, maps them through a shared trainable embedding 
table, and mean-pools each sequence into a 32-dimensional vector. 
The two vectors are concatenated into a 64-dimensional representation 
and passed through layer normalization and a multilayer perceptron 
with hidden widths of 256 and 128 and GELU activations. 
The output layer produces three logits for reliable, questionable, and no 
training evidence. The classifier is trained end to end 
with unweighted cross-entropy. 
As noted above, this baseline does not use the model's hidden
states or other internal representations. Its purpose is to
evaluate whether response reliability can be assessed without
explicit access to the model's internal states or parameters.

\subsubsection{Target Configuration with hidden-state input}

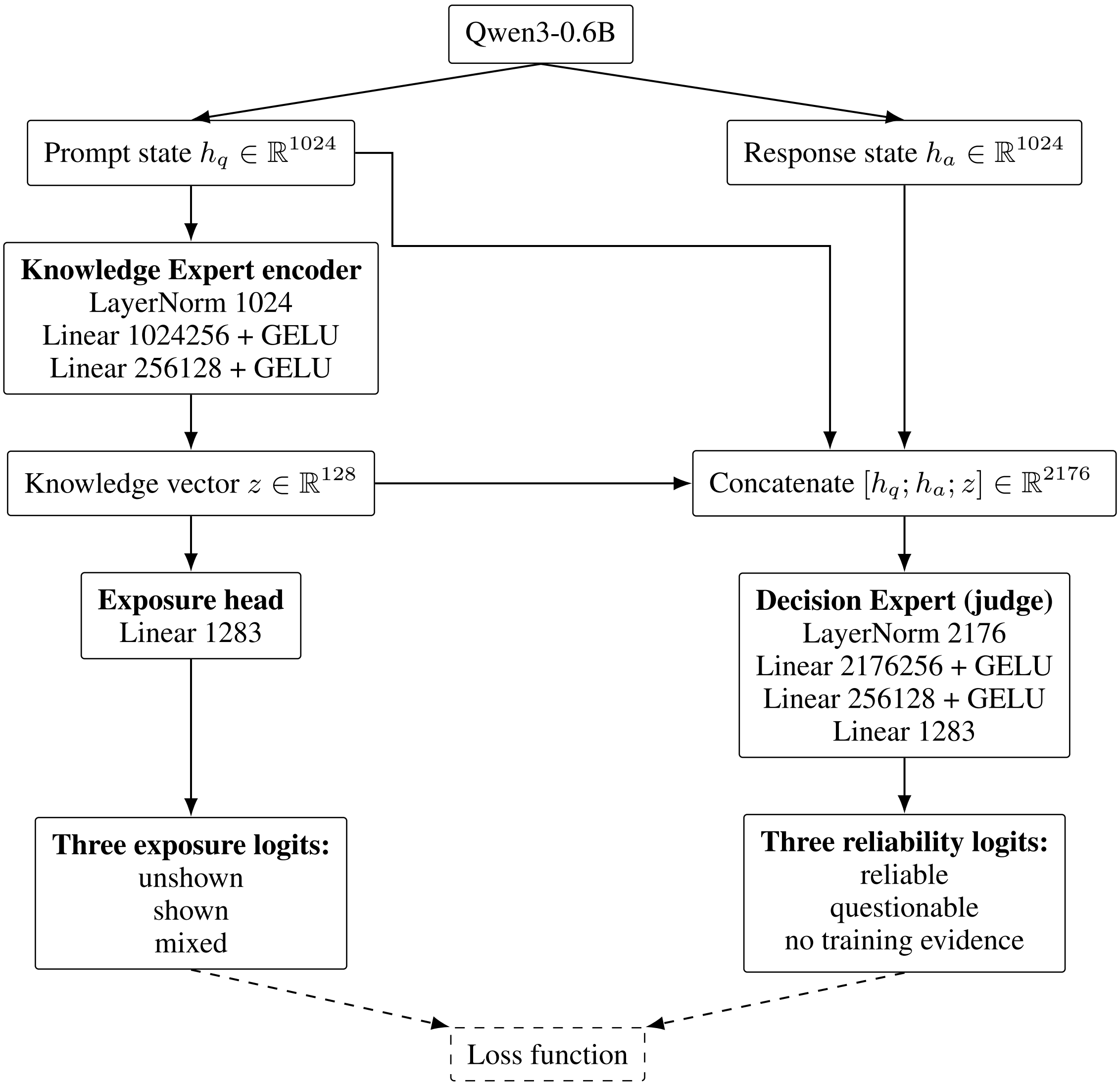
\begin{figure}[htbp]
\centering
\resizebox{0.93\linewidth}{!}{%
\begin{tikzpicture}[node distance=6mm and 10mm]
  \node[exblock] (qwen) {Qwen3-0.6B};
  \node[exblock, below left=of qwen] (hq) {Prompt state \(h_q\in\mathbb R^{1024}\)};
  \node[exblock, below right=of qwen] (ha) {Response state \(h_a\in\mathbb R^{1024}\)};

  \node[exblock, below=of hq] (knowledge) {
    \textbf{Knowledge Expert encoder}\\
    LayerNorm 1024\\
    Linear 1024256 + GELU\\
    Linear 256128 + GELU
  };
  \node[exblock, below=of knowledge] (z) {Knowledge vector \(z\in\mathbb R^{128}\)};
  \node[exblock] (concat) at (ha |- z) {
    Concatenate \([h_q;h_a;z]\in\mathbb R^{2176}\)
  };

  \node[exblock, below=of z] (exposure) {
    \textbf{Exposure head}\\
    Linear 1283
  };
  \node[exblock, below=of concat] (judge) {
    \textbf{Decision Expert (judge)}\\
    LayerNorm 2176\\
    Linear 2176256 + GELU\\
    Linear 256128 + GELU\\
    Linear 1283
  };

  \node[exblock, below=of judge] (rel) {\textbf{Three reliability logits:}\\
  reliable\\
  questionable\\
  no training evidence

  };
  \node[exblock] (exp) at (exposure |- rel) {\textbf{Three exposure logits:}\\
  unshown\\
  shown\\
  mixed
 
  };
  \path (exp.south) -- (rel.south) coordinate[midway] (outputs);
  \node[exloss, below=of outputs] (loss) {Loss function};

  \draw[exflow] (qwen.south) -- (hq.north);
  \draw[exflow] (qwen.south) -- (ha.north);
  \draw[exflow] (hq) -- (knowledge);
  \draw[exflow] (knowledge) -- (z);
  \draw[exflow] (hq.east) -- ++(4mm,0)
    -- ++(0,-10mm) -| ([xshift=-8mm]concat.north);
  \draw[exflow] (ha.south) -- (concat.north);
  \draw[exflow] (z) -- (concat);
  \draw[exflow] (z) -- (exposure);
  \draw[exflow] (concat) -- (judge);
  \draw[exflow] (exposure) -- (exp);
  \draw[exflow] (judge) -- (rel);
  \draw[extrain] (exp.south) -- (loss.north west);
  \draw[extrain] (rel.south) -- (loss.north east);
\end{tikzpicture}%
}
\caption{Target configuration. The inputs are the final-layer states at the last
formatted prompt token ($h_q$) and the last generated response token
before EOS ($h_a$). Dashed arrows indicate training-only connections
where applicable.}
\label{fig:expert-hidden}
\end{figure}

Figure~\ref{fig:expert-hidden} shows the two-expert architecture with hidden-state
input only. The model provides two
1,024-dimensional final-layer hidden states. The question state $h_q$
is read at the last token of the formatted generation prompt. Causal
masking prevents this position from attending to the response. The response
state $h_a$ is read at the last generated token before the end-of-sequence
marker and represents the question together with the generated response.

The Knowledge Expert takes $h_q$ as its sole input. It applies layer
normalization followed by affine layers of widths 256 and 128, each with a
GELU activation, to produce the knowledge vector \(z\in\mathbb{R}^{128}\).
A separate affine head maps $z$ to three exposure logits for
\emph{unshown}, \emph{shown}, and \emph{mixed}.

The Decision Expert concatenates $h_q$, $h_a$, and $z$ into a
2,176-dimensional vector. It applies layer normalization and two hidden
layers of widths 256 and 128 with GELU activations. Its output layer
produces three logits for \emph{reliable}, \emph{questionable}, and
\emph{no training evidence}. In the supervised cases, the experts are
trained jointly, with decision gradients flowing through $z$ into the
Knowledge Expert.

\subsubsection{Weight-Augmented Configuration with Model Weights and Hidden-State Inputs}

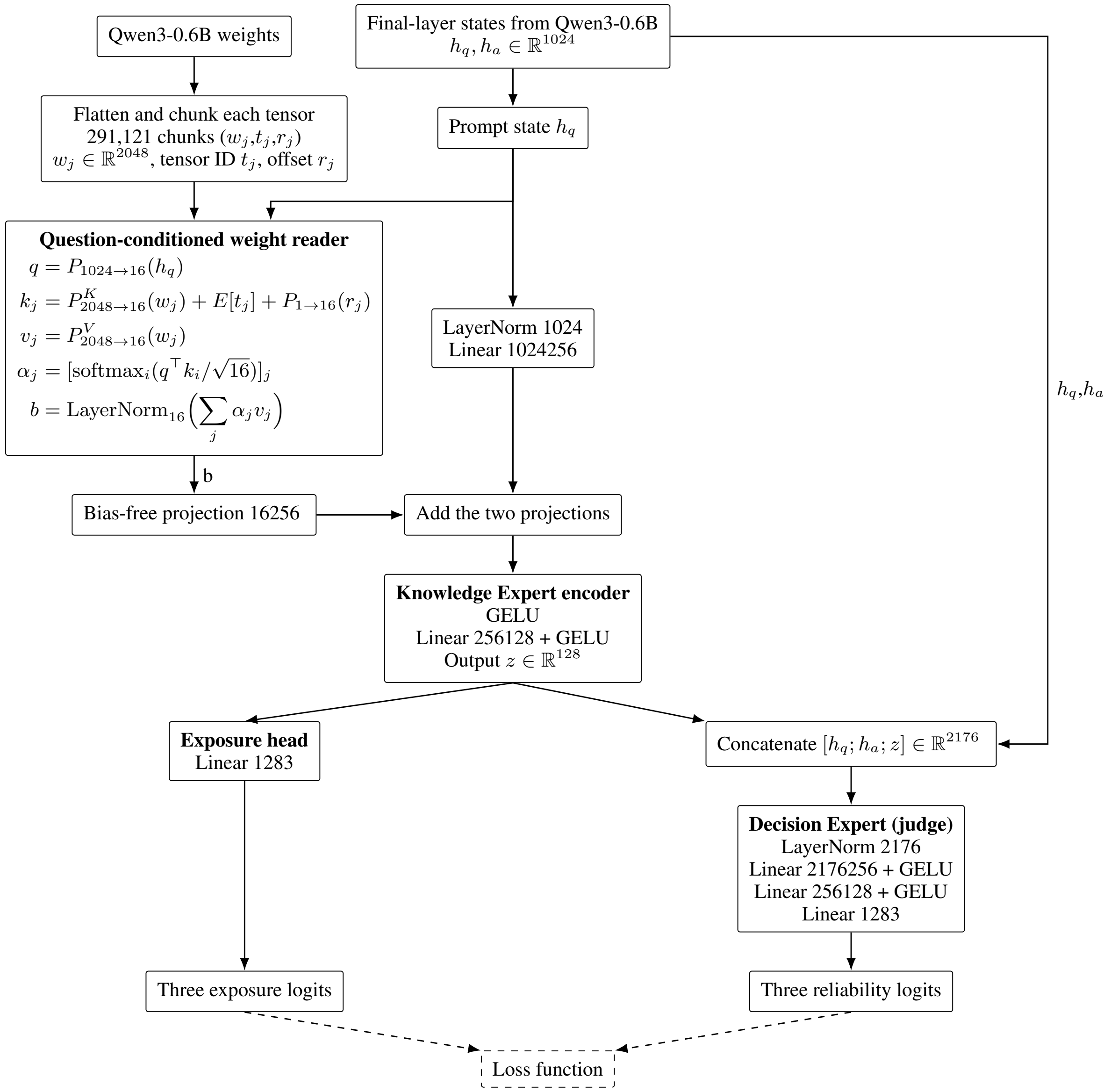
\begin{figure}[htbp]
\centering
\resizebox{0.98\linewidth}{!}{%
\begin{tikzpicture}[node distance=6mm and 10mm]
  \node[exblock] (weights) {Qwen3-0.6B weights};
  \node[exblock, right=of weights] (states) {
    Final-layer states from Qwen3-0.6B\\
    \(h_q,h_a\in\mathbb R^{1024}\)
  };

  \node[exblock, below=of weights] (chunks) {
    Flatten and chunk each tensor\\
    291,121 chunks ($w_j$,$t_j$,$r_j$)\\
    \(w_j\in\mathbb R^{2048}\), tensor ID $t_j$, offset $r_j$
  };
  \node[exblock, below=of states] (prompt) {Prompt state $h_q$};

  \node[exblock, below=of chunks] (reader) {
    \textbf{Question-conditioned weight reader}\\[2pt]
    \(\begin{aligned}
      q   &= P_{1024\to16}(h_q)\\
      k_j &= P^{K}_{2048\to16}(w_j)+E[t_j]+P_{1\to16}(r_j)\\
      v_j &= P^{V}_{2048\to16}(w_j)\\
      \alpha_j &= [\operatorname{softmax}_i(q^\top k_i/\sqrt{16})]_j\\
      b &= \operatorname{LayerNorm}_{16}\Bigl(\sum_j\alpha_jv_j\Bigr)
    \end{aligned}\)
  };
  \node[exblock] (promptproj) at (prompt |- reader) {
    LayerNorm 1024\\
    Linear 1024256
  };

  \node[exblock, below=of reader] (weightproj) {
    Bias-free projection 16256
  };
  \node[exblock] (sum) at (promptproj |- weightproj) {Add the two projections};
  \node[exblock, below=of sum] (knowledge) {
    \textbf{Knowledge Expert encoder}\\
    GELU\\
    Linear 256128 + GELU\\
    Output \(z\in\mathbb R^{128}\)
  };

  \node[exblock, below left=of knowledge] (exposure) {
    \textbf{Exposure head}\\
    Linear 1283
  };
  \node[exblock, below right=of knowledge] (concat) {
    Concatenate \([h_q;h_a;z]\in\mathbb R^{2176}\)
  };
  \node[exblock, below=of concat] (judge) {
    \textbf{Decision Expert (judge)}\\
    LayerNorm 2176\\
    Linear 2176256 + GELU\\
    Linear 256128 + GELU\\
    Linear 1283
  };

  \node[exblock, below=of judge] (rel) {Three reliability logits};
  \node[exblock] (exp) at (exposure |- rel) {Three exposure logits};
  \path (exp.south) -- (rel.south) coordinate[midway] (outputs);
  \node[exloss, below=of outputs] (loss) {Loss function};

  \draw[exflow] (weights) -- (chunks);
  \draw[exflow] (chunks) -- (reader);
  \draw[exflow] (states) -- (prompt);
  \coordinate (readerentry) at ([yshift=3mm]reader.north);
  \draw[exflow] (prompt.south)
    -- (prompt.south |- readerentry)
    -| ([xshift=12mm]reader.north);
  \draw[exflow] (prompt) -- (promptproj);
  \draw[exflow] (reader) -- node[right,font=\small] {b} (weightproj);
  \draw[exflow] (weightproj) -- (sum);
  \draw[exflow] (promptproj) -- (sum);
  \draw[exflow] (sum) -- (knowledge);
  \draw[exflow] (knowledge.south) -- (exposure.north);
  \draw[exflow] (knowledge.south) -- (concat.north west);
  \coordinate (statebypass) at ([xshift=8mm]concat.east);
  \draw[exflow] (states.east)
    -- (states.east -| statebypass)
    |- node[pos=.25,right,font=\small] {$h_q$,$h_a$} (concat.east);
  \draw[exflow] (concat) -- (judge);
  \draw[exflow] (exposure) -- (exp);
  \draw[exflow] (judge) -- (rel);
  \draw[extrain] (exp.south) -- (loss.north west);
  \draw[extrain] (rel.south) -- (loss.north east);
\end{tikzpicture}%
}
\caption{Weight-augmented configuration.}
\label{fig:expert-hidden-weights}
\end{figure}

Figure~\ref{fig:expert-hidden-weights} shows the weight-augmented
architecture, which incorporates explicit model-weight information
alongside hidden-state representations. 
The Knowledge Expert receives the question state
\(h_q\in\mathbb{R}^{1024}\) and the model's full parameters. 
Each parameter tensor is flattened and
split into chunks of 2,048 values, with zero padding at tensor boundaries.
This produces 291,121 chunks. Each chunk $w_j$ is associated with its
tensor identity $t_j$ and normalized position $r_j$ within that tensor.

The weight reader uses $h_q$ to attend over all chunks. It projects the
question state to a 16-dimensional query and each chunk to a key and
value of the same dimension. The key also incorporates a learned tensor
embedding and a projected position:
\[
q=W_qh_q,\qquad
k_j=W_kw_j+E[t_j]+W_rr_j,\qquad
v_j=W_vw_j.
\]
The reader pools the values by scaled dot-product attention:
\[
\alpha_j =
\frac{\exp(q^\top k_j/\sqrt{16})}
     {\sum_k \exp(q^\top k_k/\sqrt{16})},
\qquad
b=\operatorname{LayerNorm}\!\left(\sum_j \alpha_jv_j\right).
\]
The pooled vector $b$ is projected to 256 dimensions and added to the
256-dimensional projection of the normalized question state. The fused
representation passes through GELU, an affine layer of width 128, and a
second GELU to produce the knowledge vector $z$.

An affine exposure head maps $z$ to three exposure logits. The Decision
Expert takes the concatenated vector $[h_q;h_a;z]$ and produces three
decision logits using the same architecture as in
Figure~\ref{fig:expert-hidden}. Gradients from
both heads update the Knowledge Expert and its weight reader,
except in the random-loss configuration. The language model
parameters remain frozen.

\section{Experimental Setup}
\label{sec:experiment_setup}

The experiment was conducted as follows. First, a dataset containing
synthetic biographies and corresponding question--answer pairs was
prepared using the procedures described in Appendix~\ref{app:dataset}.
The Qwen3-0.6B model was then fine-tuned for five epochs to reproduce
information from this dataset, as described in
Appendix~\ref{app:model_training}. After each epoch, factual recall was
assessed using exact-match evaluation. Following fine-tuning, the model
was evaluated on a separate dataset using Qwen3.5-4B as an evaluator
and exact-match checks, as described in
Appendix~\ref{app:post_train_validation}. The same answer-evaluation
procedure was used in the subsequent expert-training and evaluation
stages.

Seven combinations of expert architecture and loss function were
evaluated, as summarized in Table~\ref{tab:expert_configurations}.
These comprised a Decision Expert baseline and two variants of the
two-expert architecture, each evaluated with symmetric, asymmetric,
and random-loss objectives. The random-loss configurations served as
controls: because their objectives were independent of the expert
outputs, no optimizer updates were performed.

\begin{table*}[htbp]
\centering
\small
\caption{Seven combinations of expert architecture, input, and loss
function. Here, $h_q$ and $h_a$ denote the question and response hidden
states, respectively, and $z$ denotes the knowledge vector produced
by the Knowledge Expert.}
\label{tab:expert_configurations}
\begin{tabularx}{\textwidth}{@{}c p{3.2cm} X p{3.1cm}@{}}
\toprule
\textbf{Case} & \textbf{Configuration} & \textbf{Input} & \textbf{Loss} \\
\midrule
0 & Decision Expert only
    & Question and answer text, encoded as pooled embeddings
      of hashed token unigrams and bigrams.
    & Decision cross-entropy. \\

\addlinespace
1 & Knowledge + Decision Experts
    & Knowledge Expert: $h_q$.
      Decision Expert: $[h_q; h_a; z]$.
    & Symmetric: exposure and decision cross-entropies. \\

\addlinespace
2 & Knowledge + Decision Experts
    & Knowledge Expert: $h_q$.
      Decision Expert: $[h_q; h_a; z]$.
    & Asymmetric: exposure cross-entropy, severity-weighted
      decision cross-entropy, and a false-acceptance penalty. \\

\addlinespace
3 & Knowledge + Decision Experts
    & Knowledge Expert: $h_q$.
      Decision Expert: $[h_q; h_a; z]$.
    & Random-loss control; no optimizer updates. \\

\addlinespace
4 & Knowledge + Decision Experts with model weights
    & Knowledge Expert: $h_q$ and attention over
      291,121 chunks of frozen language model parameters.
      Decision Expert: $[h_q; h_a; z]$.
    & Symmetric: exposure and decision cross-entropies. \\

\addlinespace
5 & Knowledge + Decision Experts with model weights
    & Knowledge Expert: $h_q$ and attention over
      291,121 chunks of frozen language model parameters.
      Decision Expert: $[h_q; h_a; z]$.
    & Asymmetric: exposure cross-entropy, severity-weighted
      decision cross-entropy, and a false-acceptance penalty. \\

\addlinespace
6 & Knowledge + Decision Experts with model weights
    & Knowledge Expert: $h_q$ and attention over
      291,121 chunks of frozen language model parameters.
      Decision Expert: $[h_q; h_a; z]$.
    & Random-loss control; no optimizer updates. \\
\bottomrule
\end{tabularx}
\end{table*}

For each configuration, confusion matrices, accuracy, and F1 scores
were computed over the complete evaluation set and separately for
each question group described in
Appendix~\ref{app:post_train_validation}.

\section{Results}
\label{sec:results}

Qwen3-0.6B acquired the biography information during fine-tuning.
Exact-match accuracy on the 384 sampled training questions increased
from 92.97\% after the first epoch to 100.00\% after the fifth,
while mean training loss decreased from 0.129342 to 0.000983
(Table~\ref{tab:training_performance}). At epoch five, the model
correctly answered all 119 open-recall, four full-biography, and
261 yes/no verification questions in this sample. The separate
post-training evaluation panel contained 2,000 questions from the
fine-tuning cohort and 2,000 from the held-out biography cohort
(Table~\ref{tab:evaluation_question_distribution_aftertune}).

Table~\ref{tab:expert-best-combined} summarizes expert performance.
Without a knowledge vector, the text-based Decision Expert in
Case~0 achieved 54.65\% accuracy and 53.48\% macro-F1. Its
question-type macro-F1 scores ranged from 51.21\% for open recall
to 57.36\% for full-biography recall.

\begin{table*}[htbp]
\centering
\small
\setlength{\tabcolsep}{5pt}
\renewcommand{\arraystretch}{1.2}
\caption{Best overall and question-type test results for the
Knowledge Expert (KE) and Decision Expert (DE) on the
2,000-question panel. Question types are described in
Appendix~\ref{app:post_train_validation}.
Overall cells report accuracy and macro-F1; question-type cells
report errors/questions and macro-F1. Bold indicates the highest
macro-F1 for each expert within each evaluation column.}
\label{tab:expert-best-combined}
\begin{tabular}{@{}clcccc@{}}
\toprule
& & \textbf{Overall}
& \textbf{Open recall}
& \textbf{Full biography}
& \textbf{Yes/no verification} \\
\textbf{Case} & \textbf{Expert}
& \textbf{Acc., macro-F1}
& \textbf{Errors/$n$, macro-F1}
& \textbf{Errors/$n$, macro-F1}
& \textbf{Errors/$n$, macro-F1} \\
\midrule

0 & KE & \textemdash & \textemdash & \textemdash & \textemdash \\
  & DE & 54.65\%, 53.48\%
       & 339/688, 51.21\%
       & 11/28, 57.36\%
       & 560/1284, 55.58\% \\
\midrule

1 & KE & 68.35\%, \textbf{67.39\%}
       & 366/688, \textbf{45.21\%}
       & 12/28, \textbf{56.91\%}
       & 247/1284, 79.88\% \\
  & DE & 74.25\%, 73.86\%
       & 253/688, 65.02\%
       & 6/28, 79.94\%
       & 230/1284, \textbf{81.83\%} \\
\midrule

2 & KE & 67.10\%, 65.82\%
       & 372/688, 43.30\%
       & 14/28, 50.21\%
       & 271/1284, 77.82\% \\
  & DE & 73.85\%, 74.56\%
       & 256/688, 63.88\%
       & 7/28, 75.10\%
       & 244/1284, 80.74\% \\
\midrule

3 & KE & 28.75\%, 25.80\%
       & 527/688, 16.36\%
       & 21/28, 17.68\%
       & 877/1284, 27.23\% \\
  & DE & 39.90\%, 19.01\%
       & 420/688, 18.69\%
       & 18/28, 17.54\%
       & 764/1284, 19.22\% \\
\midrule

4 & KE & 68.55\%, 66.62\%
       & 372/688, 41.80\%
       & 14/28, 49.58\%
       & 228/1284, \textbf{81.11\%} \\
  & DE & 74.30\%, 74.74\%
       & 238/688, 67.31\%
       & 8/28, 74.20\%
       & 268/1284, 78.20\% \\
\midrule

5 & KE & 67.35\%, 65.12\%
       & 372/688, 43.73\%
       & 12/28, 56.28\%
       & 264/1284, 78.41\% \\
  & DE & 75.30\%, \textbf{76.11\%}
       & 228/688, \textbf{68.80\%}
       & 4/28, \textbf{85.56\%}
       & 238/1284, 81.05\% \\
\midrule

6 & KE & 28.65\%, 25.61\%
       & 522/688, 18.26\%
       & 20/28, 22.76\%
       & 885/1284, 25.56\% \\
  & DE & 39.90\%, 19.01\%
       & 420/688, 18.69\%
       & 18/28, 17.54\%
       & 764/1284, 19.22\% \\
\bottomrule
\end{tabular}

\end{table*}

The hidden-state two-expert architectures in Cases~1 and~2
achieved Decision Expert accuracies of 74.25\% and 73.85\%,
with macro-F1 scores of 73.86\% and 74.56\%, respectively.
The Knowledge Expert achieved overall macro-F1 scores of
67.39\% and 65.82\%. Its strongest question-type results were
on yes/no verification (79.88\% and 77.82\%), followed by
full-biography recall (56.91\% and 50.21\%) and open recall
(45.21\% and 43.30\%). Exposure classification was more accurate
for shown and unshown biographies than for mixed cases:
Case~1 achieved 70.1\%, 74.5\%, and 52.5\%, respectively,
and Case~2 achieved 69.5\%, 73.5\%, and 49.5\%
(Table~\ref{tab:knowledge-exposure-accuracy}).


\begin{longtable}{@{}llrrr@{}}
\caption{Knowledge Expert test accuracy by question type and fine-tuning
exposure.}
\label{tab:knowledge-exposure-accuracy}\\
\toprule
Case index & Question type & Shown & Unshown & Mixed \\
\midrule
\endfirsthead
\toprule
Case index & Question type & Shown & Unshown & Mixed \\
\midrule
\endhead
\bottomrule
\endlastfoot
1 & All         & 561/800 (70.1\%) & 596/800 (74.5\%) & 210/400 (52.5\%) \\
1 & Open recall & 137/269 (50.9\%) & 142/269 (52.8\%) & 43/150 (28.7\%) \\
1 & Biography   & 8/11 (72.7\%)    & 3/11 (27.3\%)    & 5/6 (83.3\%) \\
1 & Yes/no      & 434/520 (83.5\%) & 435/520 (83.7\%) & 168/244 (68.9\%) \\
\addlinespace

2 & All         & 556/800 (69.5\%) & 588/800 (73.5\%) & 198/400 (49.5\%) \\
2 & Open recall & 119/269 (44.2\%) & 162/269 (60.2\%) & 35/150 (23.3\%) \\
2 & Biography   & 5/11 (45.5\%)    & 6/11 (54.5\%)    & 3/6 (50.0\%) \\
2 & Yes/no      & 431/520 (82.9\%) & 421/520 (81.0\%) & 161/244 (66.0\%) \\
\addlinespace

3 & All         & 251/800 (31.4\%) & 27/800 (3.4\%)   & 297/400 (74.2\%) \\
3 & Open recall & 0/269 (0.0\%)    & 22/269 (8.2\%)   & 139/150 (92.7\%) \\
3 & Biography   & 0/11 (0.0\%)     & 1/11 (9.1\%)     & 6/6 (100.0\%) \\
3 & Yes/no      & 251/520 (48.3\%) & 4/520 (0.8\%)    & 152/244 (62.3\%) \\
\addlinespace

4 & All         & 593/800 (74.1\%) & 596/800 (74.5\%) & 182/400 (45.5\%) \\
4 & Open recall & 117/269 (43.5\%) & 170/269 (63.2\%) & 29/150 (19.3\%) \\
4 & Biography   & 6/11 (54.5\%)    & 5/11 (45.5\%)    & 3/6 (50.0\%) \\
4 & Yes/no      & 464/520 (89.2\%) & 429/520 (82.5\%) & 163/244 (66.8\%) \\
\addlinespace

5 & All         & 549/800 (68.6\%) & 623/800 (77.9\%) & 175/400 (43.8\%) \\
5 & Open recall & 116/269 (43.1\%) & 162/269 (60.2\%) & 38/150 (25.3\%) \\
5 & Biography   & 7/11 (63.6\%)    & 6/11 (54.5\%)    & 3/6 (50.0\%) \\
5 & Yes/no      & 453/520 (87.1\%) & 401/520 (77.1\%) & 166/244 (68.0\%) \\
\addlinespace

6 & All         & 255/800 (31.9\%) & 34/800 (4.2\%)   & 284/400 (71.0\%) \\
6 & Open recall & 2/269 (0.7\%)    & 32/269 (11.9\%)  & 132/150 (88.0\%) \\
6 & Biography   & 0/11 (0.0\%)     & 2/11 (18.2\%)    & 6/6 (100.0\%) \\
6 & Yes/no      & 253/520 (48.7\%) & 0/520 (0.0\%)    & 146/244 (59.8\%) \\
\end{longtable}

The asymmetric objective in Case~2 changed the distribution
of Decision Expert errors relative to the symmetric objective
in Case~1. Accuracy on shown cases decreased from 87.1\% to
73.2\%, while accuracy on unshown and mixed cases increased
from 64.8\% to 72.9\% and from 67.5\% to 77.0\%, respectively
(Table~\ref{tab:decision-exposure-accuracy}). For open recall,
shown-case accuracy decreased from 60.2\% to 35.7\%, whereas
unshown-case accuracy increased from 56.5\% to 75.8\%.

The random-loss control in Case~3 performed substantially
below both supervised configurations, with macro-F1 scores
of 25.80\% for the Knowledge Expert and 19.01\% for the
Decision Expert (Table~\ref{tab:expert-best-combined}).
No parameter updates were performed. The Decision Expert's
results were consistent with a constant reliable-class
prediction: accuracy was 99.8\% on shown cases and 0.0\%
on both unshown and mixed cases
(Table~\ref{tab:decision-exposure-accuracy}).
The Knowledge Expert instead showed a strong bias toward
mixed exposure, achieving 74.2\% accuracy on mixed cases
but only 31.4\% and 3.4\% on shown and unshown cases,
respectively (Table~\ref{tab:knowledge-exposure-accuracy}).

Adding model-weight input in Cases~4 and~5 produced modest
overall Decision Expert gains relative to Cases~1 and~2:
macro-F1 increased from 73.86\% to 74.74\% under symmetric
loss and from 74.56\% to 76.11\% under asymmetric loss.
Knowledge Expert macro-F1 decreased from 67.39\% to 66.62\%
and from 65.82\% to 65.12\%, respectively.
Question-type improvements were uneven: Case~5 achieved
the highest Decision Expert macro-F1 for open recall
(68.80\%) and full-biography recall (85.56\%), while Case~1
retained the highest yes/no verification score (81.83\%).
Full-biography results were based on only 28 questions.
The weight-augmented random-loss control, Case~6, retained
the same Decision Expert accuracy and macro-F1 as Case~3
(39.90\% and 19.01\%;
Table~\ref{tab:expert-best-combined}).

\begin{longtable}{@{}llrrr@{}}
\caption{Decision Expert test accuracy by question type and
fine-tuning exposure. Accuracy measures agreement with the decision
label within each exposure group.}
\label{tab:decision-exposure-accuracy}\\
\toprule
Case index & Question type & Shown & Unshown & Mixed \\
\midrule
\endfirsthead
\toprule
Case index & Question type & Shown & Unshown & Mixed \\
\midrule
\endhead
\bottomrule
\endlastfoot
0 & All         & 420/800 (52.5\%) & 473/800 (59.1\%) & 200/400 (50.0\%) \\
0 & Open recall & 139/269 (51.7\%) & 148/269 (55.0\%) & 62/150 (41.3\%) \\
0 & Biography   & 8/11 (72.7\%)    & 8/11 (72.7\%)    & 1/6 (16.7\%) \\
0 & Yes/no      & 290/520 (55.8\%) & 288/520 (55.4\%) & 146/244 (59.8\%) \\
\addlinespace

1 & All         & 697/800 (87.1\%) & 518/800 (64.8\%) & 270/400 (67.5\%) \\
1 & Open recall & 162/269 (60.2\%) & 152/269 (56.5\%) & 121/150 (80.7\%) \\
1 & Biography   & 11/11 (100.0\%)  & 6/11 (54.5\%)    & 5/6 (83.3\%) \\
1 & Yes/no      & 469/520 (90.2\%) & 390/520 (75.0\%) & 195/244 (79.9\%) \\
\addlinespace

2 & All         & 586/800 (73.2\%) & 583/800 (72.9\%) & 308/400 (77.0\%) \\
2 & Open recall & 96/269 (35.7\%)  & 204/269 (75.8\%) & 132/150 (88.0\%) \\
2 & Biography   & 5/11 (45.5\%)    & 11/11 (100.0\%)  & 5/6 (83.3\%) \\
2 & Yes/no      & 424/520 (81.5\%) & 425/520 (81.7\%) & 191/244 (78.3\%) \\
\addlinespace

3 & All         & 798/800 (99.8\%) & 0/800 (0.0\%)    & 0/400 (0.0\%) \\
3 & Open recall & 268/269 (99.6\%) & 0/269 (0.0\%)    & 0/150 (0.0\%) \\
3 & Biography   & 10/11 (90.9\%)   & 0/11 (0.0\%)     & 0/6 (0.0\%) \\
3 & Yes/no      & 520/520 (100.0\%)& 0/520 (0.0\%)    & 0/244 (0.0\%) \\
\addlinespace

4 & All         & 620/800 (77.5\%) & 553/800 (69.1\%) & 313/400 (78.2\%) \\
4 & Open recall & 150/269 (55.8\%) & 169/269 (62.8\%) & 131/150 (87.3\%) \\
4 & Biography   & 8/11 (72.7\%)    & 7/11 (63.6\%)    & 5/6 (83.3\%) \\
4 & Yes/no      & 462/520 (88.8\%) & 377/520 (72.5\%) & 177/244 (72.5\%) \\
\addlinespace

5 & All         & 639/800 (79.9\%) & 533/800 (66.6\%) & 334/400 (83.5\%) \\
5 & Open recall & 144/269 (53.5\%) & 170/269 (63.2\%) & 146/150 (97.3\%) \\
5 & Biography   & 10/11 (90.9\%)   & 10/11 (90.9\%)   & 4/6 (66.7\%) \\
5 & Yes/no      & 427/520 (82.1\%) & 426/520 (81.9\%) & 193/244 (79.1\%) \\
\addlinespace

6 & All         & 798/800 (99.8\%) & 0/800 (0.0\%)    & 0/400 (0.0\%) \\
6 & Open recall & 268/269 (99.6\%) & 0/269 (0.0\%)    & 0/150 (0.0\%) \\
6 & Biography   & 10/11 (90.9\%)   & 0/11 (0.0\%)     & 0/6 (0.0\%) \\
6 & Yes/no      & 520/520 (100.0\%)& 0/520 (0.0\%)    & 0/244 (0.0\%) \\
\end{longtable}

\section{Discussion and Conclusion}
\label{sec:discussion}

The results support the feasibility of a two-expert architecture
that constructs a question-conditioned knowledge vector and passes
it to a response-level Decision Expert. The hidden-state
configurations achieved 73.85--74.25\% decision accuracy and
73.86--74.56\% macro-F1
(Table~\ref{tab:expert-best-combined}). These accuracies fall
within the 71--83\% range reported by
\citet{azaria-mitchell-2023-internal} for binary truthfulness
classification. This comparison provides context rather than
evidence of equivalent performance: the present experiment uses
a smaller model, different data, and three target classes.
Moreover, although both two-expert configurations outperformed
Case~0, that baseline uses text features rather than hidden
states. A matched hidden-state baseline without the knowledge
vector is therefore needed to isolate the vector's incremental
contribution.

A distinguishing feature of this experiment is its control over
factual exposure. The synthetic biographies provide a record of
which canonical biographies were included in fine-tuning, whereas
the pretraining exposure of facts in conventional question-answering
benchmarks is often uncertain. The complete recall of the sampled
training questions after five epochs confirms that the model
learned the evaluated training examples
(Table~\ref{tab:training_performance}). Nevertheless, exposure
is an operational proxy for knowledge, rather than direct proof
that every relevant association is retained or accessible.
Conversely, excluding a canonical biography from fine-tuning
does not establish that all its component names and values were
unseen: some could occur in verification questions as false
premises.

The contribution of this study extends beyond the number of
knowledge categories. \citet{yin-etal-2024-benchmarking} examine
knowledge boundaries through prompt sensitivity and answer
elicitation, while
\citet{xiao2025kscopeframeworkcharacterizingknowledge}
characterize the consistency and correctness of answer
distributions through repeated sampling. The present study
instead examines whether a learned, question-conditioned
representation can predict controlled fine-tuning exposure
and support a separate response-level decision. Its experimental
setting distinguishes shown, unshown, and mixed exposure,
allowing exposure prediction to be evaluated separately from
answer assessment. The comparisons further examine the
additional value of explicit model-weight input and the effect
of asymmetric error costs. 

The Knowledge Expert identified exposure more successfully for
shown and unshown biographies than for mixed cases. In Cases~1
and~2, accuracy was 69.5--70.1\% for shown cases and
73.5--74.5\% for unshown cases, compared with 49.5--52.5\%
for mixed cases
(Table~\ref{tab:knowledge-exposure-accuracy}). This difference
indicates a particular difficulty in representing partial
exposure within a single question-conditioned vector.
It is distinct from the effect of question format:
yes/no verification produced the highest Knowledge Expert
macro-F1, whereas open recall produced the lowest
(Table~\ref{tab:expert-best-combined}). Mixed exposure and
question type are separate dimensions, so the results do not
establish that every mixed question is harder than every
full-biography or verification question.

The two experts also differed in performance. Overall exposure
accuracy ranged from 67.10\% to 68.55\% across the supervised
configurations, below the Decision Expert's
73.85--75.30\%. These figures describe overall performance;
Knowledge Expert macro-F1 on open recall was substantially
lower, at 41.80--45.21\%. Thus, identifying exposure from the
question alone was less successful than classifying response
outcomes after observing the generated response. This distinction
complements the separation between internally recoverable and
externally expressed knowledge examined by
\citet{gekhman2025insideouthiddenfactualknowledge}.
However, it does not demonstrate the reverse of their finding:
the experts receive different information and predict different
labels. The present results show that exposure prediction and
response assessment have different empirical difficulty,
rather than establishing a direct measure of stored knowledge.

Explicit access to model weights provided limited additional
benefit under the tested architecture and training settings.
Compared with their hidden-state counterparts, Cases~4 and~5
increased Decision Expert macro-F1 by 0.88 and 1.55 percentage
points, respectively, while reducing Knowledge Expert macro-F1
by 0.77 and 0.70 points
(Table~\ref{tab:expert-best-combined}). Improvements also
varied by question type. The hidden-state architecture therefore
provides a reasonable starting point without the additional
weight reader. 

The loss comparison demonstrates that similar aggregate
performance can accompany different error distributions.
The asymmetric objective increases the cost of decision errors
according to estimated severity and additionally penalizes
assigning high reliable-class probability to examples labelled
otherwise. In the implemented experiment, severity was
approximated through lexical reference matching and required
claim counts; contradiction and unsupported-claim counts were
not separately activated. Relative to Case~1, Case~2 increased
Decision Expert macro-F1 only slightly, from 73.86\% to
74.56\%, but shifted accuracy from shown cases toward unshown
and mixed cases
(Table~\ref{tab:decision-exposure-accuracy}). This suggests
that cost-sensitive training can adjust the trade-off between
types of error. Demonstrating more conservative acceptance,
however, requires class-specific false-acceptance rates and
precision--recall analysis; the reported group accuracies alone
do not establish probability calibration or a universally
safer decision rule.

The random-loss controls provide a complementary comparison.
Because their objectives were independent of expert outputs,
they performed no parameter updates. Their Decision Expert
results were consistent with a constant reliable-class
prediction, yielding 39.90\% accuracy but only 19.01\%
macro-F1. The Knowledge Expert instead showed a strong bias
toward mixed exposure
(Tables~\ref{tab:expert-best-combined}
and~\ref{tab:knowledge-exposure-accuracy}). The supervised
experts therefore learned discriminative signals absent from
these fixed-parameter controls. This comparison does not,
by itself, establish that the signals exclusively represent
factual knowledge rather than question or answer regularities.

Such regularities are relevant to practical deployment.
\citet{villuri2026internalprobesbeatreading} show that
answer-surface features and output-logit information can explain
some apparent advantages of internal probes. Case~0 similarly
demonstrates that response assessment is possible using text
features alone. The proposed hidden-state experts use compact
multilayer perceptrons and a 128-dimensional knowledge vector,
making them plausible candidates for further efficiency
analysis. In contrast, the weight-augmented architecture
attends over 291,121 parameter chunks and should not be assumed
to have comparable cost. Neither runtime overhead nor the
minimum sufficient expert size was measured here.

Training auxiliary experts alongside language-model fine-tuning
is a possible extension. The required hidden states could be
obtained from language-model forward passes, but the current
experts were trained against a frozen model. Concurrent training
would introduce changing representations and require suitable
exposure and response labels, potentially including generated
answers and external evaluation. Its computational cost and
stability must therefore be measured before concluding that
it introduces no substantial bottleneck. If these requirements
can be met, auxiliary experts could provide an additional
reliability signal for task-specific small models. Extending
from factual recall to whether a model can perform a trained
task remains a separate empirical question.

Finally, these experiments used a single seed and a limited
set of expert architectures and optimization settings. The
reported best test scores are retrospective selections, and
full-biography results rely on only 28 questions. The findings
therefore establish preliminary architectural feasibility
rather than optimality, statistical robustness, or calibrated
confidence. Repeated-seed experiments, matched feature
ablations, independently selected checkpoints, and evaluation
on new biographies, question formulations, and tasks are
needed to determine how far the observed behavior generalizes.

\section*{Data Availability Statement}
The code, experimental logs, and results associated with this
study are available in the GitHub repository\footnote{\url{https://github.com/susurofu/two-expert-architecture}}. Due to size
limitations, the repository does not include the training and
evaluation datasets or cached model-generated responses. Readers
interested in accessing these materials may open an issue in the
repository or contact the author by email.

\section*{Declaration of Generative AI Use}
ChatGPT was used for (1) assistance with coding, debugging, and
code review; (2) proofreading, summarizing text fragments, and
refining the wording of the manuscript; and (3) assistance with
\LaTeX{} formatting. Qwen3.5-4B was employed as part of the
evaluation pipeline, as described in the manuscript. Google
Scholar's integrated AI tools were also used to assist with
the literature review.

\bibliography{custom}

@inproceedings{NEURIPS2025_90043ebd,
 author = {Zhao, Eric and Awasthi, Pranjal and Haghtalab, Nika},
 booktitle = {Advances in Neural Information Processing Systems},
 doi = {10.52202/085713-3335},
 editor = {D. Belgrave and C. Zhang and H. Lin and R. Pascanu and P. Koniusz and M. Ghassemi and N. Chen},
 pages = {99780--99809},
 publisher = {Curran Associates, Inc.},
 title = {From Style to Facts: Mapping the Boundaries of Knowledge Injection with Finetuning},
 url = {https://proceedings.neurips.cc/paper_files/paper/2025/file/90043ebd68500f9efe84fedf860a64f3-Paper-Conference.pdf},
 volume = {38, Main Conference},
 year = {2025}
}

@misc{meng2023locatingeditingfactualassociations,
      title={Locating and Editing Factual Associations in GPT}, 
      author={Kevin Meng and David Bau and Alex Andonian and Yonatan Belinkov},
      year={2023},
      eprint={2202.05262},
      archivePrefix={arXiv},
      primaryClass={cs.CL},
      url={https://arxiv.org/abs/2202.05262}, 
}

@misc{todd2024functionvectorslargelanguage,
      title={Function Vectors in Large Language Models}, 
      author={Eric Todd and Millicent L. Li and Arnab Sen Sharma and Aaron Mueller and Byron C. Wallace and David Bau},
      year={2024},
      eprint={2310.15213},
      archivePrefix={arXiv},
      primaryClass={cs.CL},
      url={https://arxiv.org/abs/2310.15213}, 
}

@misc{kadavath2022languagemodelsmostlyknow,
      title={Language Models (Mostly) Know What They Know}, 
      author={Saurav Kadavath and Tom Conerly and Amanda Askell and Tom Henighan and Dawn Drain and Ethan Perez and Nicholas Schiefer and Zac Hatfield-Dodds and Nova DasSarma and Eli Tran-Johnson and Scott Johnston and Sheer El-Showk and Andy Jones and Nelson Elhage and Tristan Hume and Anna Chen and Yuntao Bai and Sam Bowman and Stanislav Fort and Deep Ganguli and Danny Hernandez and Josh Jacobson and Jackson Kernion and Shauna Kravec and Liane Lovitt and Kamal Ndousse and Catherine Olsson and Sam Ringer and Dario Amodei and Tom Brown and Jack Clark and Nicholas Joseph and Ben Mann and Sam McCandlish and Chris Olah and Jared Kaplan},
      year={2022},
      eprint={2207.05221},
      archivePrefix={arXiv},
      primaryClass={cs.CL},
      url={https://arxiv.org/abs/2207.05221}, 
}

@inproceedings{azaria-mitchell-2023-internal,
    title = "The Internal State of an {LLM} Knows When It{'}s Lying",
    author = "Azaria, Amos  and
      Mitchell, Tom",
    editor = "Bouamor, Houda  and
      Pino, Juan  and
      Bali, Kalika",
    booktitle = "Findings of the {A}ssociation for {C}omputational {L}inguistics: {EMNLP} 2023",
    month = dec,
    year = "2023",
    address = "Singapore",
    publisher = "Association for Computational Linguistics",
    url = "https://aclanthology.org/2023.findings-emnlp.68/",
    doi = "10.18653/v1/2023.findings-emnlp.68",
    pages = "967--976"
}

@misc{burns2024discoveringlatentknowledgelanguage,
      title={Discovering Latent Knowledge in Language Models Without Supervision}, 
      author={Collin Burns and Haotian Ye and Dan Klein and Jacob Steinhardt},
      year={2024},
      eprint={2212.03827},
      archivePrefix={arXiv},
      primaryClass={cs.CL},
      url={https://arxiv.org/abs/2212.03827}, 
}

@article{farquhar2024detecting,
  title={Detecting hallucinations in large language models using semantic entropy},
  author={Farquhar, Sebastian and Kossen, Jannik and Kuhn, Lorenz and Gal, Yarin},
  journal={Nature},
  volume={630},
  number={8017},
  pages={625--630},
  year={2024},
  publisher={Nature Publishing Group UK, London}
}

@misc{kossen2024semanticentropyprobesrobust,
      title={Semantic Entropy Probes: Robust and Cheap Hallucination Detection in LLMs}, 
      author={Jannik Kossen and Jiatong Han and Muhammed Razzak and Lisa Schut and Shreshth Malik and Yarin Gal},
      year={2024},
      eprint={2406.15927},
      archivePrefix={arXiv},
      primaryClass={cs.CL},
      url={https://arxiv.org/abs/2406.15927}, 
}

@misc{orgad2025llmsknowshowintrinsic,
      title={LLMs Know More Than They Show: On the Intrinsic Representation of LLM Hallucinations}, 
      author={Hadas Orgad and Michael Toker and Zorik Gekhman and Roi Reichart and Idan Szpektor and Hadas Kotek and Yonatan Belinkov},
      year={2025},
      eprint={2410.02707},
      archivePrefix={arXiv},
      primaryClass={cs.CL},
      url={https://arxiv.org/abs/2410.02707}, 
}

@misc{ferrando2025iknowentityknowledge,
      title={Do I Know This Entity? Knowledge Awareness and Hallucinations in Language Models}, 
      author={Javier Ferrando and Oscar Obeso and Senthooran Rajamanoharan and Neel Nanda},
      year={2025},
      eprint={2411.14257},
      archivePrefix={arXiv},
      primaryClass={cs.CL},
      url={https://arxiv.org/abs/2411.14257}, 
}

@misc{gekhman2025insideouthiddenfactualknowledge,
      title={Inside-Out: Hidden Factual Knowledge in LLMs}, 
      author={Zorik Gekhman and Eyal Ben David and Hadas Orgad and Eran Ofek and Yonatan Belinkov and Idan Szpektor and Jonathan Herzig and Roi Reichart},
      year={2025},
      eprint={2503.15299},
      archivePrefix={arXiv},
      primaryClass={cs.CL},
      url={https://arxiv.org/abs/2503.15299}, 
}

@inproceedings{ni-etal-2025-towards,
    title = "Towards Fully Exploiting {LLM} Internal States to Enhance Knowledge Boundary Perception",
    author = "Ni, Shiyu  and
      Bi, Keping  and
      Guo, Jiafeng  and
      Yu, Lulu  and
      Bi, Baolong  and
      Cheng, Xueqi",
    editor = "Che, Wanxiang  and
      Nabende, Joyce  and
      Shutova, Ekaterina  and
      Pilehvar, Mohammad Taher",
    booktitle = "Proceedings of the 63rd Annual Meeting of the {A}ssociation for {C}omputational {L}inguistics (Volume 1: Long Papers)",
    month = jul,
    year = "2025",
    address = "Vienna, Austria",
    publisher = "Association for Computational Linguistics",
    url = "https://aclanthology.org/2025.acl-long.1184/",
    doi = "10.18653/v1/2025.acl-long.1184",
    pages = "24315--24329",
    ISBN = "979-8-89176-251-0"
}

@misc{cencerrado2026answerneededpredictingllm,
      title={No Answer Needed: Predicting LLM Answer Accuracy from Question-Only Linear Probes}, 
      author={Iván Vicente Moreno Cencerrado and Arnau Padrés Masdemont and Anton Gonzalvez Hawthorne and David Demitri Africa and Lorenzo Pacchiardi},
      year={2026},
      eprint={2509.10625},
      archivePrefix={arXiv},
      primaryClass={cs.CL},
      url={https://arxiv.org/abs/2509.10625}, 
}

@misc{chen2026queryleveluncertaintylargelanguage,
      title={Query-Level Uncertainty in Large Language Models}, 
      author={Lihu Chen and Gerard de Melo and Fabian M. Suchanek and Gaël Varoquaux},
      year={2026},
      eprint={2506.09669},
      archivePrefix={arXiv},
      primaryClass={cs.CL},
      url={https://arxiv.org/abs/2506.09669}, 
}

@inproceedings{ashuach-etal-2026-masked,
    title = "Masked by Consensus: Disentangling Privileged Knowledge in {LLM} Correctness",
    author = "Ashuach, Tomer  and
      Gretz, Shai  and
      Katz, Yoav  and
      Belinkov, Yonatan  and
      Ein-Dor, Liat",
    editor = "Liakata, Maria  and
      Moreira, Viviane P.  and
      Zhang, Jiajun  and
      Jurgens, David",
    booktitle = "Proceedings of the 64th Annual Meeting of the {A}ssociation for {C}omputational {L}inguistics (Volume 1: Long Papers)",
    month = jul,
    year = "2026",
    address = "San Diego, California, United States",
    publisher = "Association for Computational Linguistics",
    url = "https://aclanthology.org/2026.acl-long.483/",
    doi = "10.18653/v1/2026.acl-long.483",
    pages = "10577--10596",
    ISBN = "979-8-89176-390-6"
}

@misc{wang2026gradeprobingknowledgegaps,
      title={GRADE: Probing Knowledge Gaps in LLMs through Gradient Subspace Dynamics}, 
      author={Yujing Wang and Yuanbang Liang and Yukun Lai and Hainan Zhang and Hanqi Yan},
      year={2026},
      eprint={2604.02830},
      archivePrefix={arXiv},
      primaryClass={cs.CL},
      url={https://arxiv.org/abs/2604.02830}, 
}

@misc{marks2024geometrytruthemergentlinear,
      title={The Geometry of Truth: Emergent Linear Structure in Large Language Model Representations of True/False Datasets}, 
      author={Samuel Marks and Max Tegmark},
      year={2024},
      eprint={2310.06824},
      archivePrefix={arXiv},
      primaryClass={cs.AI},
      url={https://arxiv.org/abs/2310.06824}, 
}

@inproceedings{huang-etal-2026-uncovering,
    title = "Uncovering the Knowledge Boundaries and Mechanistic ``Knob'' of {LLM}s",
    author = "Huang, Zihao  and
      Mao, Rui  and
      Jiang, Yong  and
      Cambria, Erik",
    editor = "Lee, Hung-Yi  and
      Dev, Sunipa  and
      Li, Jessy  and
      Elliott, Desmond",
    booktitle = "Findings of the {A}ssociation for {C}omputational {L}inguistics: {EMNLP} 2026",
    month = oct,
    year = "2026",
    address = "Budapest, Hungary",
    publisher = "Association for Computational Linguistics",
    url = "https://aclanthology.org/2026.findings-emnlp.1262/",
    doi = "10.18653/2026.findings-emnlp.1262",
    pages = "26679--26694",
    ISBN = "979-8-89176-462-0"
}

@misc{villuri2026internalprobesbeatreading,
      title={When Do Internal Probes Beat Reading the Answer? Miscalibrated Readouts and Behavior-Concealed Knowledge in Language Models}, 
      author={Gnaneswar Villuri and Hashmath Shaik and Alex Doboli},
      year={2026},
      eprint={2609.04582},
      archivePrefix={arXiv},
      primaryClass={cs.CL},
      url={https://arxiv.org/abs/2609.04582}, 
}

@article{2017arXivtriviaqa,
       author = {{Joshi}, Mandar and {Choi}, Eunsol and {Weld},
                 Daniel and {Zettlemoyer}, Luke},
        title = "{triviaqa: A Large Scale Distantly Supervised Challenge Dataset for Reading Comprehension}",
      journal = {arXiv e-prints},
         year = 2017,
          eid = {arXiv:1705.03551},
        pages = {arXiv:1705.03551},
archivePrefix = {arXiv},
       eprint = {1705.03551},
}

@article{sentence_transformers2024natural_questions,
    title = "Natural Questions: A Benchmark for Question Answering Research",
    author = "Kwiatkowski, Tom  and
      Palomaki, Jennimaria  and
      Redfield, Olivia  and
      Collins, Michael  and
      Parikh, Ankur  and
      Alberti, Chris  and
      Epstein, Danielle  and
      Polosukhin, Illia  and
      Devlin, Jacob  and
      Lee, Kenton  and
      Toutanova, Kristina  and
      Jones, Llion  and
      Kelcey, Matthew  and
      Chang, Ming-Wei  and
      Dai, Andrew M.  and
      Uszkoreit, Jakob  and
      Le, Quoc  and
      Petrov, Slav",
    editor = "Lee, Lillian  and
      Johnson, Mark  and
      Roark, Brian  and
      Nenkova, Ani",
    journal = "Transactions of the Association for Computational Linguistics",
    volume = "7",
    year = "2019",
    address = "Cambridge, MA",
    publisher = "MIT Press",
    url = "https://aclanthology.org/Q19-1026/",
    doi = "10.1162/tacl_a_00276",
    pages = "452--466"
}

@inproceedings{amayuelas-etal-2024-knowledge,
    title = "Knowledge of Knowledge: Exploring Known-Unknowns Uncertainty with Large Language Models",
    author = "Amayuelas, Alfonso  and
      Wong, Kyle  and
      Pan, Liangming  and
      Chen, Wenhu  and
      Wang, William Yang",
    editor = "Ku, Lun-Wei  and
      Martins, Andre  and
      Srikumar, Vivek",
    booktitle = "Findings of the {A}ssociation for {C}omputational {L}inguistics: {ACL} 2024",
    month = aug,
    year = "2024",
    address = "Bangkok, Thailand",
    publisher = "Association for Computational Linguistics",
    url = "https://aclanthology.org/2024.findings-acl.383/",
    doi = "10.18653/v1/2024.findings-acl.383",
    pages = "6416--6432"
}

@inproceedings{NEURIPS2024_6fdf57c7,
 author = {Chang, Hoyeon and Park, Jinho and Ye, Seonghyeon and Yang, Sohee and Seo, Youngkyung and Chang, Du-Seong and Seo, Minjoon},
 booktitle = {Advances in Neural Information Processing Systems},
 doi = {10.52202/079017-1939},
 editor = {A. Globerson and L. Mackey and D. Belgrave and A. Fan and U. Paquet and J. Tomczak and C. Zhang},
 pages = {60626--60668},
 publisher = {Curran Associates, Inc.},
 title = {How Do Large Language Models Acquire Factual Knowledge During Pretraining?},
 url = {https://proceedings.neurips.cc/paper_files/paper/2024/file/6fdf57c71bc1f1ee29014b8dc52e723f-Paper-Conference.pdf},
 volume = {37},
 year = {2024}
}

@inproceedings{ou-etal-2025-llms,
    title = "How Do {LLM}s Acquire New Knowledge? A Knowledge Circuits Perspective on Continual Pre-Training",
    author = "Ou, Yixin  and
      Yao, Yunzhi  and
      Zhang, Ningyu  and
      Jin, Hui  and
      Sun, Jiacheng  and
      Deng, Shumin  and
      Li, Zhenguo  and
      Chen, Huajun",
    editor = "Che, Wanxiang  and
      Nabende, Joyce  and
      Shutova, Ekaterina  and
      Pilehvar, Mohammad Taher",
    booktitle = "Findings of the {A}ssociation for {C}omputational {L}inguistics: {ACL} 2025",
    month = jul,
    year = "2025",
    address = "Vienna, Austria",
    publisher = "Association for Computational Linguistics",
    url = "https://aclanthology.org/2025.findings-acl.1021/",
    doi = "10.18653/v1/2025.findings-acl.1021",
    pages = "19889--19913",
    ISBN = "979-8-89176-256-5"
}

@misc{zucchet2025languagemodelslearnfacts,
      title={How do language models learn facts? Dynamics, curricula and hallucinations}, 
      author={Nicolas Zucchet and Jörg Bornschein and Stephanie Chan and Andrew Lampinen and Razvan Pascanu and Soham De},
      year={2025},
      eprint={2503.21676},
      archivePrefix={arXiv},
      primaryClass={cs.CL},
      url={https://arxiv.org/abs/2503.21676}, 
}

@inproceedings{dai-etal-2022-knowledge,
    title = "Knowledge Neurons in Pretrained Transformers",
    author = "Dai, Damai  and
      Dong, Li  and
      Hao, Yaru  and
      Sui, Zhifang  and
      Chang, Baobao  and
      Wei, Furu",
    editor = "Muresan, Smaranda  and
      Nakov, Preslav  and
      Villavicencio, Aline",
    booktitle = "Proceedings of the 60th Annual Meeting of the {A}ssociation for {C}omputational {L}inguistics (Volume 1: Long Papers)",
    month = may,
    year = "2022",
    address = "Dublin, Ireland",
    publisher = "Association for Computational Linguistics",
    url = "https://aclanthology.org/2022.acl-long.581/",
    doi = "10.18653/v1/2022.acl-long.581",
    pages = "8493--8502"
}

@misc{xiao2025kscopeframeworkcharacterizingknowledge,
      title={KScope: A Framework for Characterizing the Knowledge Status of Language Models}, 
      author={Yuxin Xiao and Shan Chen and Jack Gallifant and Danielle Bitterman and Thomas Hartvigsen and Marzyeh Ghassemi},
      year={2025},
      eprint={2506.07458},
      archivePrefix={arXiv},
      primaryClass={cs.CL},
      url={https://arxiv.org/abs/2506.07458}, 
}

@inproceedings{ishii-etal-2025-fine,
    title = "Fine-grained Confidence Estimation for Spurious Correctness Detection in Large Language Models",
    author = "Ishii, Ai  and
      Inoue, Naoya  and
      Suzuki, Hisami  and
      Sekine, Satoshi",
    editor = "Inui, Kentaro  and
      Sakti, Sakriani  and
      Wang, Haofen  and
      Wong, Derek F.  and
      Bhattacharyya, Pushpak  and
      Banerjee, Biplab  and
      Ekbal, Asif  and
      Chakraborty, Tanmoy  and
      Singh, Dhirendra Pratap",
    booktitle = "Proceedings of the 14th International Joint Conference on Natural Language Processing and the 4th Conference of the {A}sia-Pacific Chapter of the {A}ssociation for {C}omputational {L}inguistics",
    month = dec,
    year = "2025",
    address = "Mumbai, India",
    publisher = "The Asian Federation of Natural Language Processing and The Association for Computational Linguistics",
    url = "https://aclanthology.org/2025.ijcnlp-long.68/",
    doi = "10.18653/v1/2025.ijcnlp-long.68",
    pages = "1238--1257",
    ISBN = "979-8-89176-298-5"
}

@misc{qwen3technicalreport,
      title={Qwen3 Technical Report}, 
      author={Qwen Team},
      year={2025},
      eprint={2505.09388},
      archivePrefix={arXiv},
      primaryClass={cs.CL},
      url={https://arxiv.org/abs/2505.09388}, 
}

@misc{li2025knowledgeboundarylargelanguage,
      title={Knowledge Boundary of Large Language Models: A Survey}, 
      author={Moxin Li and Yong Zhao and Wenxuan Zhang and Shuaiyi Li and Wenya Xie and See-Kiong Ng and Tat-Seng Chua and Yang Deng},
      year={2025},
      eprint={2412.12472},
      archivePrefix={arXiv},
      primaryClass={cs.CL},
      url={https://arxiv.org/abs/2412.12472}, 
}

@inproceedings{yin-etal-2024-benchmarking,
    title = "Benchmarking Knowledge Boundary for Large Language Models: A Different Perspective on Model Evaluation",
    author = "Yin, Xunjian  and
      Zhang, Xu  and
      Ruan, Jie  and
      Wan, Xiaojun",
    editor = "Ku, Lun-Wei  and
      Martins, Andre  and
      Srikumar, Vivek",
    booktitle = "Proceedings of the 62nd Annual Meeting of the {A}ssociation for {C}omputational {L}inguistics (Volume 1: Long Papers)",
    month = aug,
    year = "2024",
    address = "Bangkok, Thailand",
    publisher = "Association for Computational Linguistics",
    url = "https://aclanthology.org/2024.acl-long.124/",
    doi = "10.18653/v1/2024.acl-long.124",
    pages = "2270--2286"
}

\newpage
\appendix
\section{Synthetic biography dataset}
\label{app:dataset}

The dataset contains 2,000 fictional people generated 
by the script. The generator forms 2,000 unique names by taking 
the Cartesian product of 40 first names and 50 surnames, shuffles 
the roster using seed 731, and groups adjacent entries into 1,000 
two-person households. The two members of each household share a city, 
street address, and residence start year, and each names the 
other as their housemate.

Each biography record contains the person's name and five
biographical attributes, which are combined into a textual description.

\begin{table*}[htbp]
\centering
\caption{Structure of the synthetic personal information dataset.}
\label{tab:dataset_structure}
\begin{tabularx}{\textwidth}{l X l}
\toprule
\textbf{Field} & \textbf{Meaning} & \textbf{Example} \\
\midrule
\texttt{person} 
    & Full name identifying the subject 
    & Ada Dalewood \\
\texttt{birth\_year} 
    & Year of birth 
    & 1961 \\
\texttt{home\_city} 
    & City of current residence 
    & Amberhaven \\
\texttt{home\_street} 
    & Street address, including the house number 
    & 1 Willow Lane \\
\texttt{residence\_since} 
    & Year the person moved into the current residence 
    & 1995 \\
\texttt{housemate} 
    & Full name of the other household member 
    & Samuel Yarrow \\
\bottomrule
\end{tabularx}
\end{table*}

The biographies follow the format illustrated below.
\begin{quotation}
  Ada Dalewood was born in 1961. Ada Dalewood 
  lives in Amberhaven. Ada Dalewood lives on 
  1 Willow Lane. Ada Dalewood has lived at the current 
  residence since 1995. Ada Dalewood shares a home with Samuel Yarrow.
\end{quotation}

For each roster entry, the generator constructs questions about 
the named person alone and about that person paired with the person 
two positions later in the roster, wrapping around at the end. For each 
subject configuration, it enumerates 26 attribute scopes: five 
single-attribute scopes, ten two-attribute scopes, ten three-attribute 
scopes, and one full-biography scope covering all five attributes. 
In a two-person question, the same attributes are requested for both 
people, and the answer follows the order of the names in the question.

\begin{table*}[htbp]
\centering
\caption{Question types used for model fine-tuning and factual recall evaluation.}
\label{tab:question_types}
\begin{tabularx}{\textwidth}{p{3.2cm} X X}
\toprule
\textbf{Question type} & \textbf{Example} & \textbf{Expected answer} \\
\midrule

Single-attribute recall 
    & In what year was Ada Dalewood born?
    & Ada Dalewood was born in 1961. \\

\addlinespace
Multi-attribute recall
    & Give the birth year, city of residence for Ada Dalewood.
    & Ada Dalewood was born in 1961. Ada Dalewood lives in Amberhaven. \\

\addlinespace
Full-biography recall
    & Provide information about Ada Dalewood.
    & The five-sentence biography shown above. \\

\addlinespace
Verification of true claims
    & Is all of the following information correct? Ada Dalewood 
    was born in 1961.
    & Yes. \\

\addlinespace
Verification with a false claim
    & Is all of the following information correct? Ada Dalewood 
    was born in 1962.
    & No. \\

\bottomrule
\end{tabularx}
\end{table*}

\section{Model Fine-Tuning for Memorization of Biographical Facts}
\label{app:model_training}

Qwen3-0.6B received 155,844 question–answer pairs in each of the five completed
epochs. Each epoch used the same pairs, shuffled into a new order. 
Every question was paired with its correct answer. The pairs were 
drawn from the first 1,000 synthetic biographies, with both subjects 
required to be in that subset for two-person questions. The distribution of
questions in the training dataset is summarized in 
Table~\ref{tab:training_question_distribution}.

\begin{table*}[htbp]
\centering
\caption{Distribution of training questions by question type across five training epochs.}
\label{tab:training_question_distribution}
\begin{tabularx}{\textwidth}{X r r r r}
\toprule
\textbf{Question type} 
    & \shortstack{\textbf{One person}\\\textbf{per epoch}}
    & \shortstack{\textbf{Two people}\\\textbf{per epoch}}
    & \shortstack{\textbf{Total}\\\textbf{per epoch}}
    & \shortstack{\textbf{Presentations}\\\textbf{over 5 epochs}} \\
\midrule

Open recall of 1--3 attributes per person
    & 25,000 & 24,950 & 49,950 & 249,750 \\

\addlinespace
Full-biography recall
    & 1,000 & 998 & 1,998 & 9,990 \\

\addlinespace
Verification with gold answer \texttt{Yes.}
    & 26,000 & 25,948 & 51,948 & 259,740 \\

\addlinespace
Verification with gold answer \texttt{No.}
    & 26,000 & 25,948 & 51,948 & 259,740 \\

\midrule
\textbf{Total}
    & \textbf{78,000}
    & \textbf{77,844}
    & \textbf{155,844}
    & \textbf{779,220} \\
\bottomrule
\end{tabularx}
\end{table*}

The base model underwent full-parameter supervised fine-tuning.
Training was performed on an NVIDIA RTX 4060 Ti GPU.
The setup is summarized in Table~\ref{tab:training_parameters}.

\begin{table*}[htbp]
\centering
\caption{Model training parameters and optimization settings.}
\label{tab:training_parameters}
\begin{tabularx}{\textwidth}{p{3.5cm} X}
\toprule
\textbf{Setting} & \textbf{Value} \\
\midrule

Loss
    & Answer-only causal cross-entropy, averaged over supervised tokens.
    Prompt and padding tokens were masked. The answer and end-of-message
    token were supervised. \\

\addlinespace
Optimizer
    & AdamW with weight decay of 0.01. \\

\addlinespace
Learning rate
    & $2 \times 10^{-5}$, with 3\% warmup and linear decay over
    the configured 10-epoch schedule. \\

\addlinespace
Batch size
    & 4 examples per microbatch, 4 gradient accumulation steps,
    nominally 16 examples per optimizer update. \\

\addlinespace
Gradient clipping
    & Maximum gradient norm of 1.0. \\

\addlinespace
Precision
    & BF16 autocast with FP32 model and optimizer weights. \\

\addlinespace
Gradient checkpointing
    & Enabled. \\

\addlinespace
Context length
    & 256-token limit. Actual formatted examples ranged from
    26 to 166 tokens and were kept in full. \\

\addlinespace
Seed
    & 731. \\

\bottomrule
\end{tabularx}
\end{table*}

The model generated answers greedily on the same 384 
sampled training questions after each epoch. The table 
gives exact matches by question type and mean training cross-entropy 
per supervised token. This check measures recall on training examples.

\begin{table*}[htbp]
\centering
\caption{Training-set recall by question type and training loss across epochs.}
\label{tab:training_performance}
\begin{tabularx}{\textwidth}{c c c c X r}
\toprule
\textbf{Epoch}
    & \shortstack{\textbf{Open}\\\textbf{recall}}
    & \shortstack{\textbf{Full}\\\textbf{biography}}
    & \shortstack{\textbf{Yes/no}\\\textbf{verification}}
    & \shortstack{\textbf{Overall}\\\textbf{exact match}}
    & \shortstack{\textbf{Mean}\\\textbf{training loss}} \\
\midrule

1 & 109/119 & 4/4 & 244/261 & 357/384 (92.97\%) & 0.129342 \\

2 & 118/119 & 4/4 & 258/261 & 380/384 (98.96\%) & 0.005179 \\

3 & 117/119 & 4/4 & 261/261 & 382/384 (99.48\%) & 0.002617 \\

4 & 116/119 & 3/4 & 260/261 & 379/384 (98.70\%) & 0.001646 \\

5 & 119/119 & 4/4 & 261/261 
    & \textbf{384/384 (100.00\%)} & 0.000983 \\

\bottomrule
\end{tabularx}
\end{table*}

\section{Evaluation of Biographical Fact Retrieval}
\label{app:post_train_validation}

After fine-tuning the model to memorize synthetic biographies
and retrieve their factual content, its ability to correctly
answer biography-related questions was evaluated.

\subsection{Evaluation questions}

The panel contained 2,000 questions from the fine-tuning 
set and 2,000 questions about people whose canonical biographies 
were excluded from fine-tuning. Questions that paired a person 
from each cohort were excluded. Some names or values from the 
held-out cohort could still appear as false premises in training 
verification questions.

\begin{table*}[htbp]
\centering
\caption{Distribution of evaluation questions across the 
fine-tuning and held-out biography cohorts.}
\label{tab:evaluation_question_distribution_aftertune}
\begin{tabularx}{\textwidth}{X r r}
\toprule
\textbf{Question type}
    & \shortstack{\textbf{Fine-tuning}\\\textbf{cohort}}
    & \shortstack{\textbf{Held-out biography}\\\textbf{cohort}} \\
\midrule

Open recall
    & 1,068 & 1,108 \\

\addlinespace
Full-biography recall
    & 308 & 303 \\

\addlinespace
Yes/no verification
    & 624 & 589 \\

\midrule
\textbf{Total per model}
    & \textbf{2,000} & \textbf{2,000} \\
\bottomrule
\end{tabularx}
\end{table*}

\section{Expert training and reported checkpoints}
\label{app:expert-training}

\paragraph{Data and supervision.}
The expert dataset comprised 10,000 training, 2,000 development, and
2,000 test question--reply pairs. People were split by household, and
each split contained 40\% shown, 40\% unshown, and 20\% mixed-exposure
questions. The frozen epoch-5 Qwen3-0.6B checkpoint generated one greedy
reply per question. Reply matching required normalized equality with
the reference and end-of-sequence completion. Specifically, the
comparison used case-folded word and number tokens after Unicode
normalization. The Knowledge Expert was supervised by the
three-way exposure label. The Decision Expert was supervised to
predict \emph{reliable} for matching shown replies,
\emph{questionable} for mixed questions or nonmatching shown replies,
and \emph{no training evidence} for unshown questions. These targets
are distinct from the semantic judgments in Appendix~\ref{app:post_train_validation}.

\paragraph{Optimization.}
Case~0 trained the text-only Decision Expert using unweighted decision
cross-entropy. The supervised two-expert configurations trained the
Knowledge and Decision Experts jointly with a single optimizer and
backward pass. The decision gradient flowed through the knowledge
vector into the Knowledge Expert and, where present, its trainable
weight reader. Qwen's parameters remained frozen. Cases~1 and~4
used the symmetric loss, and Cases~2 and~5 used the asymmetric loss.
The penalty counts in the asymmetric runs were derived from
the normalized reply match. Cases~3 and~6 used independent uniform
random losses with seed~9317 and made zero parameter updates.

\begin{table}[htbp]
\centering
\small
\caption{Common expert training settings.}
\label{tab:selected-expert-training-settings}
\begin{tabular}{@{}ll@{}}
\toprule
Setting & Value \\
\midrule
Initialization and shuffle seed & 731 \\
Optimizer for supervised runs & AdamW \\
AdamW coefficients & $\beta_1=0.9$, $\beta_2=0.999$, $\epsilon=10^{-8}$ \\
Weight decay & 0.01 \\
Maximum gradient norm & 1.0 \\
Batch size and shuffle & 256 pairs, reshuffled each epoch \\
Optimizer updates per supervised epoch & 40 \\
Learning rate schedule & Constant within each run \\
Epoch budget and patience, joint cases & 10 epochs, patience 5 \\
Epoch budgets for reported Case~0 runs & 20 and 40, with patience equal to the budget \\
\bottomrule
\end{tabular}
\end{table}

\paragraph{Checkpoint selection.}
Within a supervised run, the development-selected checkpoint minimized
decision cross-entropy for Case~0, or the sum of exposure and
unweighted decision cross-entropy for the joint cases. Table~2
reports the higher test macro-F1 among the development-selected
and final checkpoints across the available runs, separately for each
expert and question group. Thus, its maxima are retrospective
comparisons. Table~\ref{tab:reported-expert-checkpoints} lists the
actual checkpoints behind those cells. In particular, Case~0's reported
overall score comes from the final epoch~40, whereas its minimum
development loss occurred at epoch~5. The highest overall Decision
Expert result came from Case~5 at learning rate $5\times10^{-4}$,
at the development-selected epoch~6 after 240 optimizer updates.
Its test accuracy was 75.30\% and its macro-F1 was 76.11\%.
The highest overall Knowledge Expert macro-F1 was 67.39\%,
from Case~1 at learning rate $10^{-3}$ and epoch~9.

\begin{table}[htbp]
\centering
\small
\setlength{\tabcolsep}{4pt}
\caption{Learning rate and epoch for the reported results in Table~2.
An asterisk marks a reported final epoch distinct from the
development-selected epoch.}
\label{tab:reported-expert-checkpoints}
\begin{tabular}{@{}ccllll@{}}
\toprule
Case & Expert & Overall & Open recall & Biography & Yes/no \\
\midrule
0 & DE & \(0.0005,40^{*}\) & \(0.0005,40^{*}\) & \(0.0005,40^{*}\) & \(0.0005,20^{*}\) \\
\addlinespace
1 & KE & \(0.001,9\) & \(0.0005,8\) & \(0.0005,8\) & \(0.001,9\) \\
1 & DE & \(0.001,9\) & \(0.0003,6\) & \(0.001,9\) & \(0.001,9\) \\
\addlinespace
2 & KE & \(0.001,9\) & \(0.001,9\) & \(0.0005,10^{*}\) & \(0.001,9\) \\
2 & DE & \(0.0005,6\) & \(0.0005,6\) & \(0.0003/0.0005,6\) & \(0.001,9\) \\
\addlinespace
3 & KE, DE & \multicolumn{4}{l@{}}{Epoch~1 (identical to epoch~6)} \\
\addlinespace
4 & KE & \(0.001,9\) & \(0.0005,6\) & \(0.0005,6\) & \(0.001,9\) \\
4 & DE & \(0.0005,6\) & \(0.0005,6\) & \(0.0005,6\) & \(0.0005,6\) \\
\addlinespace
5 & KE & \(0.001,9\) & \(0.001,9\) & \(0.0005,6\) & \(0.0005,6\) \\
5 & DE & \(0.0005,6\) & \(0.0005,6\) & \(0.001,9\) & \(0.001,9\) \\
\addlinespace
6 & KE, DE & \multicolumn{4}{l@{}}{Epoch~1 (identical to epoch~6)} \\
\bottomrule
\end{tabular}

\medskip
\parbox{0.94\linewidth}{\footnotesize
KE denotes the Knowledge Expert and DE the Decision Expert. Case~2's
full-biography Decision Expert scores tied at both listed learning
rates. The random-loss control runs had a nominal learning rate of
0.001 and completed six epochs with unchanged parameters.}
\end{table}


\end{document}